\documentclass[twoside]{article}

\usepackage[preprint]{aistats2027}
\usepackage[american]{babel}

\usepackage[numbers,sort&compress]{natbib}  

\usepackage{graphicx}  
\usepackage{booktabs}  
\usepackage{xcolor}
\usepackage{colortbl}  
\usepackage{multirow}
\usepackage{url}
\usepackage{algorithm}
\usepackage{algpseudocode}
\usepackage{placeins}
\usepackage{capt-of}  
\usepackage[breaklinks]{hyperref}

\usepackage{amsmath}
\usepackage{amsthm}
\usepackage{mathtools}
\usepackage{amsfonts,bm}

\newcommand{\xx}{\mathbf{x}}

\newcommand{\btheta}{\bm{\theta}}

\newcommand{\dd}{\mathbf{d}}

\newcommand{\pd}{{p^*}}
\newcommand{\psf}{p_{\sigma}}
\newcommand{\xt}{\tilde{\xx}}

\newcommand{\pe}{p_{\mathrm{emp}}}
\newcommand{\ee}{\bm{\varepsilon}}

\newcommand{\bmm}{\mathbf{\bar m}}
\newcommand{\mm}{\mathbf{m}}

\newcommand{\dat}{\mathcal{D}}

\newcommand{\cond}{\,|\,}

\begin{document}

\runningtitle{Diffusion-free Score Matching for Exact Conditional Sampling}
\runningauthor{Diffusion-free Score Matching for Exact Conditional Sampling}

\twocolumn[

\aistatstitle{Diffusion-free Score Matching for Exact Conditional Sampling}

\aistatsauthor{ Thomas Wedenig \And Robert Peharz }

\aistatsaddress{ TU Graz \\ Graz Center for Machine Learning (GraML) \And TU Graz \\ Graz Center for Machine Learning (GraML)} ]

\begin{abstract}
The success of modern score-based generative models is largely driven by diffusion-based training, which learns a collection of time-indexed score fields corresponding to progressively noisier versions of the data distribution. While such models achieve impressive sample quality, leveraging them for sound probabilistic inference---particularly sampling from \emph{arbitrary conditionals of the model}---remains challenging. This difficulty arises because conditioning information is only available for clean data and not for the higher noise levels required to generate exact conditional samples.
In this paper, we introduce \emph{DIffusion-free SCOre matching} (DISCO), a principled approach that sidesteps the need for time-dependent score fields altogether. Our method is based on a principled objective that, while reminiscent of diffusion-based training, estimates only a \emph{single} score of the (slightly perturbed) data distribution, for which the exact conditional score is readily obtained.
In experiments, DISCO is competitive with state-of-the-art diffusion models in terms of sample quality. More importantly, DISCO yields a more faithful representation of the underlying data distribution and---crucially---enables accurate sampling from arbitrary conditionals, outperforming standard heuristic samplers. This capability opens the door to principled and flexible probabilistic reasoning with score-based models.
\end{abstract}

\section{Introduction}

Diffusion models currently define the state of the art in generative modeling, with remarkable success in image synthesis, molecular design, and protein structure generation \citep{song2020sde,watson2023novo}.
At their core, they arise from classical score matching \citep{hyvarinen2005estimation} and its denoising variant \citep{dsm}, which learn the score function rather than the density itself.
Classical score matching learns the score mainly near the data manifold, leaving low-density regions unconstrained, so that sampling trajectories are easily led astray.
Diffusion-based models \citep{sohl2015deep,song2020sde} mitigate this by progressively broadening the region where the score is learned and sampling then iteratively denoises from high to low noise.

\begin{figure*}
\centering
\includegraphics[width=\textwidth]{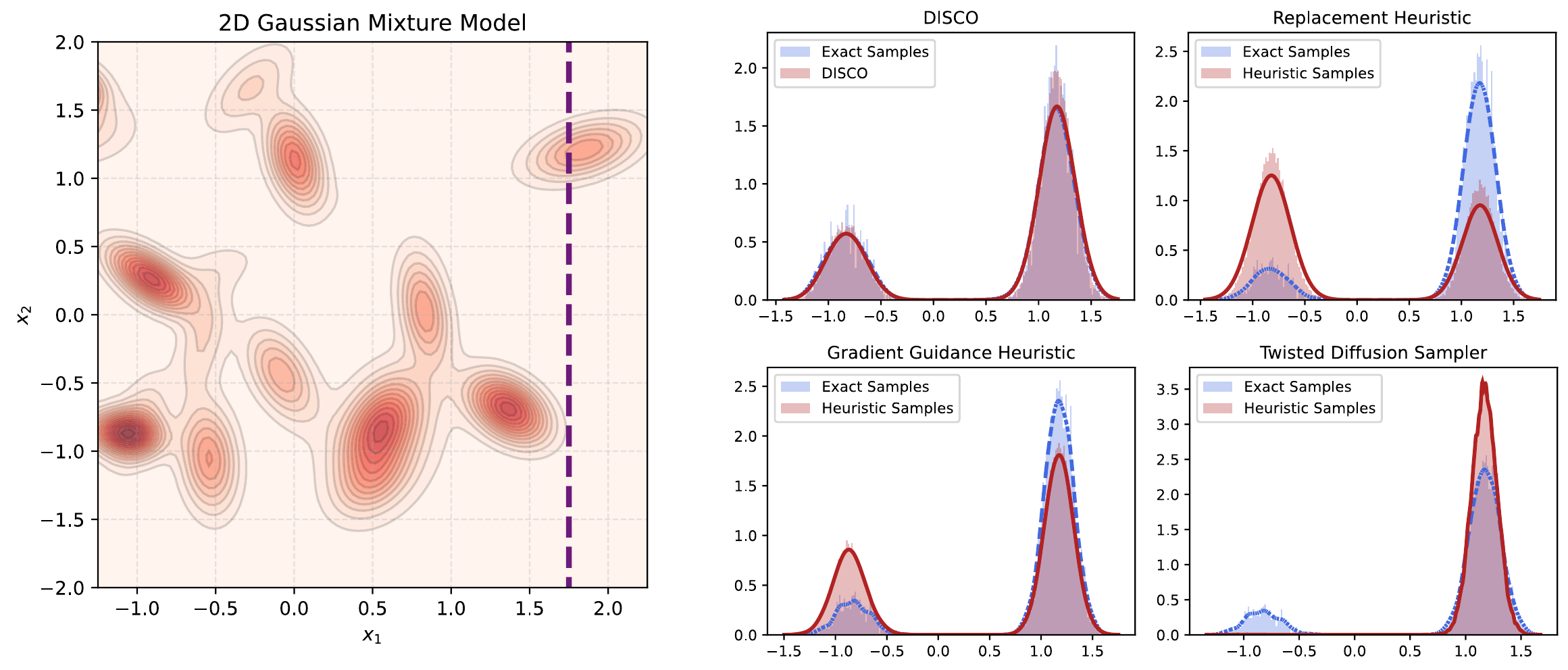}
\caption{
Conditional sampling with a diffusion model vs.\ DISCO (a \emph{single} score field) trained on a 2D GMM ($\pd$, left), for the query \mbox{$x_2 \sim p_{\mathbf{\theta}}(x_2 \cond x_1=1.75)$}.
\emph{Gradient Guidance} \citep{ho2022video}, the \emph{Replacement Heuristic} \citep{song2020sde} and the \emph{Twisted Diffusion Sampler} (TDS) \citep{wu2023practical} deviate substantially from the diffusion model's true conditional (via rejection sampling), whereas DISCO, using tempered SMC on the exact conditional score \eqref{eq:conditional_score}, matches it. Details in Section~\ref{sec:lowD} and Appendix~\ref{app:experimental_details}.
}
\label{fig:low_dim_inference}
\end{figure*}

However, the diffusion mechanism requires learning \emph{an entire family of score vector fields}, one for each noise level.
This design choice has important consequences for principled \emph{probabilistic inference}---the hallmark of probability theory as a rigorous framework for reasoning under uncertainty \citep{jaynes1995probability,pearl1988probabilistic}, which ultimately reduces to two fundamental operations: \emph{marginalization} and \emph{conditioning} \citep{ghahramani2015probabilistic}.
In the context of score-based generative models this translates into the question whether faithful \emph{samples} can be drawn from the model's marginals and conditionals.
While drawing from marginals is trivial---simply discard variables from joint samples---drawing from \emph{conditionals} requires conditioning the \emph{entire} diffusion process, which is generally intractable.
Existing approaches rely on ad-hoc heuristics \citep{song2020sde,ho2022video,kawar2022denoising} or on particle-based methods \citep{wu2023practical}, which are asymptotically exact but can be difficult to tune in practice.
As demonstrated in Figure~\ref{fig:low_dim_inference}, \emph{conditional samples of these methods are often substantially biased}---even on elementary toy problems like GMMs---revealing the brittleness of diffusion-based probabilistic reasoning.

In this work, we challenge the common assumption that expressive score-based modeling requires fitting an \emph{entire family} of noisy data scores. 
Instead, we introduce \emph{DIffusion-free SCOre matching} (DISCO), which learns a \emph{single} score field while adopting the key principle underlying diffusion models, namely broadening the region over which the score is reliably learned. Specifically, starting from a mixture of importance-weighted Fisher divergences defined through a collection of progressively noisier proposal distributions, we derive a principled score matching objective.
This objective, while reminiscent of diffusion training, \emph{fits only the (slightly perturbed) data distribution rather than a full diffusion process}, while still ensuring that the score is learned away from the data manifold. 
Conditioning then becomes conceptually trivial: fixing the observed variables in the joint score yields the exact conditional score \eqref{eq:conditional_score}, which standard Markov Chain Monte Carlo (MCMC) methods can sample.
Here, ``exact'' refers to the target: samples are \emph{asymptotically exact} draws from the conditional of the learned model.

Our experiments cover two regimes. In low dimensions (energy parameterization, Metropolis-corrected MCMC), DISCO's conditionals are asymptotically exact, whereas diffusion-based samplers are biased. On images, where we use an unconstrained score parameterization for scalability (exact only at the population optimum), DISCO's sample quality is close to that of diffusion models, and its conditionals drift the least of all tested samplers under repeated resampling.
In summary, our contributions are:

\begin{itemize}
\item We challenge the need for time-dependent diffusion and instead learn a \emph{single} joint score field, for which conditioning becomes conceptually straightforward.

\item We introduce DIffusion-free SCOre matching (DISCO), a principled objective that retains diffusion’s support-broadening benefits while learning a single score field, so that all conditionals are \emph{consistent} with one joint model.

\item In experiments, we show that DISCO is close to diffusion models in sample quality and, more importantly, yields a more faithful data representation and accurate sampling from arbitrary conditional distributions, whose consistency on images we verify by repeated resampling.
\end{itemize}
Code to reproduce all experiments is available at \url{https://anonymous.4open.science/r/disco-code/}.

\section{Background}
\label{sec:score_based_modeling}

We are given i.i.d.~samples $\{\xx^{(i)} \in \mathbb{R}^D \}_{i=1}^N$ from a data distribution $\pd$, and aim to learn a parametric model $p_{\btheta}$ that approximates $\pd$.
Score-based modeling \citep{hyvarinen2005estimation} circumvents the challenge of normalization by learning the \emph{score} rather than a density, defined as $\nabla_{\xx} \log \pd(\xx)$, which is invariant to the normalizing constant.
The idea is to use a neural network $s_{\btheta} \colon \mathbb{R}^D \to \mathbb{R}^D$ to represent the model score and minimize the \emph{Fisher divergence}:
\begin{equation}
\label{eq:fisher}
\mathcal{F}(\pd \, \| \, s_{\btheta}) := \mathbb{E}_{\xx \sim \pd}\left[ \| \nabla_{\xx} \log \pd(\xx) - s_{\btheta}(\xx) \|^2_2 \right]
\end{equation}

Rather than parameterizing the score directly, one might use an \emph{energy-based model} $E_{\btheta}(\xx)$, that defines a density via $p_{\btheta}(\xx) \coloneqq \exp(-E_{\btheta}(\xx)) / Z_{\btheta}$ and its score as $s_{\btheta}(\xx) = \nabla_{\xx} \log p_{\btheta}(\xx) = -\nabla_{\xx} E_{\btheta}(\xx)$, easily obtained via autodiff.
Such models define a proper density and admit MCMC with Metropolis--Hastings corrections such as Hamiltonian Monte Carlo (HMC) \citep{neal2011mcmc}, whereas a directly modeled score may be \emph{non-conservative} and correspond to no density \citep{chao2022investigating}---This distinction may be secondary for generation, but is essential for probabilistic reasoning \citep{jaynes1995probability,pearl1988probabilistic,ghahramani2015probabilistic}. We consider both parameterizations, despite the probabilistic advantages of the energy-based formulation.

\subsection{Denoising Score Matching}
\label{sec:dsm}

Directly minimizing the Fisher divergence in \eqref{eq:fisher} is infeasible as it depends on the unknown score, motivating alternative objectives that do not require explicit access to $\nabla_{\xx} \log \pd(\xx)$.
A widely used variant is \emph{denoising score matching} (DSM), which approximates the score of a perturbed data distribution $\psf(\xt) = \int \pd(\xx) \, q(\xt \cond \xx) \, \mathrm{d}\xx$, where $q(\xt \cond \xx) := \mathcal{N}(\xt \cond \xx, \sigma^2 I)$ is a Gaussian perturbation kernel with \emph{fixed} noise level $\sigma$.
Concretely, minimizing
\begin{equation}
\label{eq:dsm}
\mathcal{L}_{\text{DSM}}(\btheta) \coloneqq \mathbb{E}_{\pd(\xx)q(\xt \cond \xx)}\left[ \| \nabla_{\xt} \log q(\xt \cond \xx) - s_{\btheta}(\xt) \|^2_2 \right]
\end{equation}
turns out to be equivalent to minimizing Fisher divergence, as $\nabla_{\btheta}\mathcal{L}_{\text{DSM}}(\btheta) = \nabla_{\btheta} \mathcal{F}(\psf \, \| \, s_{\btheta})$ for all $\btheta$ \citep{dsm}.
This objective and its gradients can be efficiently estimated using data samples, as it only depends on the score of the perturbation kernel, given by $\nabla_{\xt} \log q(\xt \cond \xx) = (\xx - \xt) / \sigma^2$.
In \eqref{eq:dsm}, one chooses $\sigma$ to be small so that $\psf$ closely approximates the data distribution $\pd$.
Regions far from the data are thus rarely sampled, and the learned score can be inaccurate there, leading samplers astray \citep{song2019gen}.

\subsection{Diffusion Models}

Diffusion models address the limitation of DSM by learning score fields across \emph{multiple noise levels}, thereby expanding the region over which the score is reliably estimated \citep{sohl2015deep,song2020sde}.
Formally, let the clean data be denoted by $\xx \sim \pd$, and define $q_t(\xx_t \cond \xx)$ via the forward diffusion process:
\begin{equation}
\xx_t = \alpha(t) \xx + \sigma(t) \ee, \qquad \ee \sim \mathcal{N}(\mathbf{0}, I)
\end{equation}
for time $t \in [0, T]$ with some $T > 0$.
In this work, we focus primarily on the variance-exploding (VE) formulation \citep{song2020sde}, where $\alpha(t) = 1$ and only the noise scale $\sigma \colon [0, T] \to \mathbb{R}_{> 0}$ varies over time.\footnote{In our notation we assume $\sigma(0) > 0$, so that $\xx_0$ denotes a slightly noisy version of $\xx$.}
This process defines a family of progressively noisier distributions $\{p_t(\xx_t)\}_{t\in[0,T]}$ where
\begin{align}
p_t(\xx_t) &= \int q_t(\xx_t \cond \xx) \, \pd(\xx) \, \mathrm{d}\xx, \label{eq:marginal_perturbed} \\
q_t(\xx_t \cond \xx) &= \mathcal{N}(\xx_t \cond \xx, \sigma(t)^2I). \label{eq:perturbation_kernel}
\end{align}

A \emph{time-dependent} score network is then trained to approximate $s_{\btheta}(\xx_t, t) \approx \nabla_{\xx_t} \log p_t(\xx_t)$ for all $\xx_t \in \mathbb{R}^D$ and $t \in [0, T]$, by minimizing the Diffusion Model (DM) loss 
\begin{equation}
\label{eq:dm}
\mathcal{L}_{\mathrm{DM}}(\btheta) \coloneqq \mathbb{E}_{t,\xx, \xx_t} \! \left[ \lambda(t)  \|\nabla_{\xx_t} \log q_t(\xx_t \cond \xx) - s_{\btheta}(\xx_t, t) \|^2_2 \right]
\end{equation}
where $t \sim p(t)$, $\xx \sim \pd(\xx)$, and $\xx_t \sim q_t(\xx_t \cond \xx)$.
Here $p(t)$ is a distribution over $[0,T]$ and $\lambda(t)$ is a positive weighting function.
Equation~\eqref{eq:dm} can be viewed as a direct extension of \eqref{eq:dsm} to a time-indexed family of score fields $\{\nabla_{\xx_t} \log p_t(\xx_t)\}_{t\in[0,T]}$, approximated by a shared score network $s_{\btheta}(\xx_t, t)$.

After training, samples from $p_0 \approx \pd$ are generated by reversing the diffusion process, e.g., via the reverse-time SDE \citep{song2020sde} or ancestral sampling \citep{ddpm}.
Training across noise levels informs the score in low-density regions---the key advantage over DSM---but complicates \emph{probabilistic inference}, in particular conditioning, which we address next.

\newcommand{\xc}{\mathbf{x}^{c}}
\newcommand{\xu}{\mathbf{x}^{u}}
\newcommand{\supp}{\mathrm{supp}}
\newtheorem{definition}{Definition}
\newtheorem{theorem}{Theorem}

\section{Diffusion-Free Score Matching}
\label{sec:disco}

Diffusion models learn an entire family of scores, although only the distribution at $t=0$ is ultimately of interest. A key problem is that conditioning this distribution requires reasoning through the auxiliary states $\{\xx_t\}_{t \in (0,T]}$, which act as latent variables and complicate probabilistic inference \citep{ghahramani2015probabilistic}.

This difficulty with conditioning does \emph{not} arise---at least in principle---when a \emph{single} score field $\nabla_{\xx}\log p(\xx)$ is used, as in classical score matching \citep{hyvarinen2005estimation,dsm}. 
Assume that $\nabla_{\xx}\log p(\xx)$ is learned accurately, also away from the data manifold;\footnote{In theory, this should eventually occur with DSM \citep{dsm} after (prohibitive) long training time.} MCMC on this single field then yields faithful joint samples. For conditioning, we partition the variables into \textbf{u}nobserved $\xu$ and \textbf{c}onditioned $\xc$ and aim to sample $\xu \sim p(\xu \cond \xc)$.
Here $\xx = (\xu, \xc)$ with $\xu \in \mathbb{R}^{D_u}$, $\xc \in \mathbb{R}^{D_c}$, $D_u + D_c = D$; the partition is arbitrary and chosen at inference time (e.g., masked and visible pixels), so a single model answers \emph{any} conditional query---unlike autoregressive models \citep{larochelle2011neural} or triangular transport maps \citep{marzouk2016sampling}, where only some conditionals are tractable.
For a single score field, conditioning is simple and direct, as the conditional score is obtained by fixing $\xc$ in the joint score:
\begin{equation}
\label{eq:conditional_score}
\nabla_{\xu} \log p(\xu, \xc) 
=
\nabla_{\xu} \log p(\xu \mid \xc) + \underbrace{\nabla_{\xu} \log p(\xc)}_{= \bm{0}}.
\end{equation}
Since this identity is exact, MCMC can target the \emph{exact model conditional}, using the same learned score as for the joint distribution. Under standard conditions, the resulting samples are asymptotically exact; finite-compute error arises only from incomplete mixing, rather than from an approximation of the conditional target. 

In contrast, diffusion-based models maintain a family of score fields
$\{\nabla_{\xx_t} \log p_t(\xx_t)\}_{t\in[0,T]}$, and conditioning would require access to ${\nabla_{\xx_t} \log p_t(\xx_t \mid \xc)}$---conditioned on the observed \emph{clean} data $\xc$---\emph{for all noise levels} $t>0$, which is generally intractable.
While considerable effort has been devoted to approximating these conditionals \citep{song2020sde,ho2022video,kawar2022denoising,wu2023practical}, the problem remains challenging in practice (see Figure~\ref{fig:low_dim_inference}).

In this work, we challenge the assumption that diffusion-based training is strictly necessary for expressive score-based modeling. Instead, we propose to learn a \emph{single} score field, akin to classical score matching \citep{dsm}, which directly enables sampling from arbitrary conditional distributions via \eqref{eq:conditional_score}, while retaining the key idea of diffusion models, namely broadening the region over which the score is learned.
To this end, we start with a slight modification of the Fisher divergence:
\begin{definition}{q-Weighted Fisher Divergence.}
Let $\pd$ and $q$ be probability densities over $\mathbb{R}^D$ whose supports satisfy $\supp(\pd) \subseteq \supp(q)$. We define the \emph{$q$-weighted Fisher divergence} as
\begin{equation}
\label{eq:weighted_fisher}
\mathcal{F}_{q}(\pd \, \| \, s_{\btheta}) := \mathbb{E}_{\xx \sim q}\left[ \| \nabla_{\xx} \log \pd(\xx) - s_{\btheta}(\xx) \|^2_2 \right].
\end{equation}
\end{definition}
Like the Fisher divergence $\mathcal{F}$ in Equation \eqref{eq:fisher}, also $\mathcal{F}_{q}$ measures the score-mismatch between $\pd$ and the model $s_{\btheta}$, but in expectation over a \emph{proposal distribution} $q$ rather than $\pd$.
It is easy to show that $\mathcal{F}_q(\pd \, \| \, s_{\btheta}) = 0$ implies $\mathcal{F}(\pd \, \| \, s_{\btheta}) = 0$, hence $\mathcal{F}_q$ is a principled divergence.

Next, we adopt the Gaussian perturbation kernels $q_t(\xx_t \cond \xx) \coloneqq \mathcal{N}(\xx_t \cond \xx, \sigma(t)^2I)$ from \eqref{eq:perturbation_kernel} and let
\begin{equation}
\label{eq:gaussian_smoothed}
p_t(\xx_t) = \int q_t(\xx_t \cond \xx) \, \pd(\xx) \, \mathrm{d} \xx    
\end{equation}
be the corresponding Gaussian-smoothed data distributions, as in \eqref{eq:marginal_perturbed}.
We emphasize that, in contrast to diffusion models, here these smoothed distributions serve as \emph{proposal distributions} and are \emph{not} themselves targets of learning.

Note further that we adopted the parameter $t$ from diffusion models, which in our case does not signify diffusion time but just indexes the Gaussian smoothed distributions \eqref{eq:gaussian_smoothed} on various noise levels $\sigma(t) > 0$.
We further introduce, similar as in diffusion models, a prior distribution $p(t)$ over $t \in [0,T]$ and a continuous positive weighting function $\lambda(t)$. 

A key quantity in our method is the \emph{zero-level} posterior
\begin{equation}
\label{eq:posterior_perturbed}
p_{0}(\xx \cond \xx_t) = 
\frac{q_{0}(\xx_t \cond \xx) \, \pd(\xx)}{p_{0}(\xx_t)},
\end{equation}
which is the posterior distribution over clean data $\xx$ under the smallest-noise kernel $q_0$, even though $\xx_t$ may have been generated using \emph{any} noise level $t \in [0,T]$.

Now, using $p_t$ from \eqref{eq:gaussian_smoothed} as a family of proposal distributions, we introduce a \emph{weighted mixture of q-weighted Fisher divergences} as our core objective:
\begin{align}
& \mathcal{F}_{\text{mix}}(\pd \, \| \, s_{\btheta}) = 
\mathbb{E}_{p(t)} \left [\lambda(t) \, \mathcal{F}_{p_t}(\pd \, \| \, s_{\btheta}) \right] \\
& =
\mathbb{E}_{p(t)} \left [\lambda(t) \,
\mathbb{E}_{\xx_t \sim p_t}\left[ \| \nabla_{\xx_t} \log \pd(\xx_t) - s_{\btheta}(\xx_t) \|^2_2 \right]
\right]
\end{align}
Also $\mathcal{F}_{\text{mix}}$ is a principled objective, since, as $\lambda(t)$ is positive and $\mathcal{F}_{p_t}$ is non-negative, $\mathcal{F}_{\text{mix}}(\pd \, \| \, s_{\btheta}) = 0$ implies that $\mathcal{F}_{p_t}(\pd \, \| \, s_{\btheta}) = 0$ for almost all $t \in [0,T]$.

The objective $\mathcal{F}_{\text{mix}}$ requires the true data score $\nabla_{\xx} \log \pd(\xx)$, which is not available. 
Following DSM \citep{dsm}, we replace $\pd$ with the slightly perturbed $p_0$ at the lowest noise level, which closely approximates the data score for small $\sigma(0) > 0$.
With this substitution, we obtain the \emph{DIffusion-free SCOre matching} loss (DISCO loss):

\begin{theorem}
\label{theo:disco_loss}
Let 
$p(t)$ be a distribution over $[0,T]$, and $\lambda(t)$ a positive weighting function.
Further, let $p_t(\xx_t)$, $q_t(\xx_t \cond \xx)$, and $p_0(\xx \cond \xx_t)$ be defined as in \eqref{eq:marginal_perturbed}, \eqref{eq:perturbation_kernel}, and \eqref{eq:posterior_perturbed}, respectively, and let $q(t, \xx_t, \xx) \coloneqq \, p(t) \, p_t(\xx_t) \, p_0(\xx \cond \xx_t)$.
The loss $\mathcal{L}_{\mathrm{DISCO}}(\btheta)$, defined as
\begin{equation}
\label{eq:disco}
\mathbb{E}_{q(t, \xx_t, \xx)} \left[ \lambda(t) \, \|\nabla_{\xx_t} \log q_0(\xx_t \cond \xx) - s_{\btheta}(\xx_t) \|^2_2 \right]
\end{equation}
has the same parameter gradients as 
$\mathcal{F}_{\text{mix}}(p_0 \, \| \, s_{\btheta})$.
\end{theorem}

The proof can be found in Appendix \ref{app:disco_loss_proof}.
Theorem \ref{theo:disco_loss} guarantees that with sufficient capacity in $s_{\btheta}$, the global minimizer of $\mathcal{L}_{\mathrm{DISCO}}$ will learn the true score of $p_0$ in the large data limit, making DISCO a sound score matching objective.
That is, its population minimizer is $\nabla_{\xx} \log p_0(\xx)$ almost everywhere. Alternatives such as MDSM \citep{li2023learning} lack this property (Section~\ref{sec:related_work}).
$\mathcal{L}_{\mathrm{DISCO}}$ thus fits a \emph{single} score by regressing on the smallest-noise kernel score, as in DSM, while averaging over many noise levels, as in diffusion training, so that $s_{\btheta}$ is informed also far from the data.

\begin{figure}
\centering
\includegraphics[width=\linewidth]{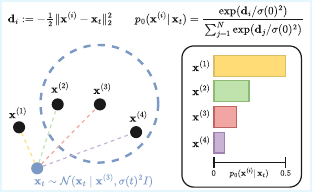}
\caption{\textbf{DISCO Training:} We pick a helper sample $\xx' = \xx^{(3)}$ from $\dat := \{\xx^{(i)}\}_{i=1}^4$ and perturb it at level $t$ to obtain $\xx_t$. Diffusion training regresses $(\xx^{(3)} - \xx_t) / \sigma(t)^2$; DISCO regresses $(\xx - \xx_t) / \sigma(0)^2$, with $\xx$ drawn from the zero-level posterior $p_0(\xx \cond \xx_t)$ \eqref{eq:posterior_perturbed}, a softmax with logits $\dd_i$ and temperature $\sigma(0)^2$. Note that $\xx^{(3)}$ need \emph{not} have high posterior probability.}
\label{fig:disco_training}
\end{figure}

\paragraph{DISCO Training.} 

As in existing score matching methods, the DISCO objective \eqref{eq:disco} is optimized by estimating the expectation and its gradients via sampling.
The key difference to diffusion training is the order of sampling: diffusion \eqref{eq:dm} draws $\xx \sim \pd$ and then $\xx_t \sim q_t(\xx_t \cond \xx)$, whereas DISCO \eqref{eq:disco} draws $\xx_t \sim p_t$ \eqref{eq:gaussian_smoothed} and then $\xx \sim p_{0}(\xx \cond \xx_t)$ \eqref{eq:posterior_perturbed}; the diffusion order would make the objective inconsistent.

Specifically, in order to sample from $q(t,\xx_t, \xx)$, we first sample $t \sim p(t)$.
As illustrated in Figure \ref{fig:disco_training}, we then sample $\xx_t \sim p_t(\xx_t)$, by first sampling some intermediate ``helper'' data sample $\xx' \sim \pd$ and then its perturbed version $\xx_t \sim q_t(\xx_t \cond \xx')$. 
The challenging part is then to sample the zero-level posterior $p_0(\xx \cond \xx_t)$.
However, as we usually have only finitely many training data points $\dat = \{\xx^{(i)}\}_{i=1}^N$, the data distribution is in fact the empirical distribution $\pd(\xx) = \pe(\xx) := \frac{1}{N} \sum_{i=1}^N \delta(\xx^{(i)} - \xx)$ where $\delta(\cdot)$ denotes the Dirac-delta. 
From Bayes' law, we obtain 
\begin{equation}
\label{eq:posterior_empirical}
p_0(\xx \cond \xx_t) = \frac{q_0(\xx_t \cond \xx) \,\pe(\xx)}{p_0(\xx_t)}
\end{equation}
which induces a probability mass function over $\dat$, as illustrated in Figure \ref{fig:disco_training}.
We then compute $p_0(\xx^{(i)} \cond \xx_t)  \propto q_0(\xx_t \cond \xx^{(i)})$ for each $\xx^{(i)} \in \dat$ and sample $\xx$ from the normalized mass function. 
For very large $\dat$, this method starts to become prohibitive, and we discuss approximately sampling from this posterior in Appendix~\ref{app:knn}. For example, one simple approximation may be to construct the posterior mass function only over a mini-batch of data.
Drawing $\xx_t \sim p_t$ and then $\xx \sim p_0(\xx \cond \xx_t)$ is simply the Monte Carlo estimator of \eqref{eq:disco}, not an inference heuristic; its only approximation is the optional mini-batch posterior, ablated in Appendix~\ref{app:minibatch_ablation}. Pseudocode is given in Algorithm~\ref{alg:disco}.

\paragraph{Masked DISCO Training.} 
Conditional sampling feeds the network inputs with clean $\xc$ and noisy $\xu$, which standard DISCO training never produces. Although the global minimizer is accurate there as well (as we confirm in low dimensions), in high dimensions the network fails to generalize to such inputs. We therefore add an auxiliary term that noises only a random \emph{subset} of coordinates, which does not change the global minimizer (Appendix~\ref{app:masked_disco_proof}).
Concretely, let $p(\mm)$ be a distribution over binary masks $\mm \in \{0, 1\}^D$, where $\mm$ is $1$ on coordinates that remain clean and $0$ on coordinates that are noised.
Let $\bmm = 1-\mm$ and write $\xx_{t,\mm}$ and $\xx_{t,\bmm}$ for the clean and noised coordinates of $\xx_t$ (and analogously $\xx_{\mm}$ and $\xx_{\bmm}$ for $\xx$), and define
$p_t(\xx_{t, \bmm},\xx_\mm)=\!\int q_t(\xx_{t, \bmm} \cond \xx_{\bmm})\,\pd(\xx_{\bmm},\xx_\mm)\,\mathrm{d}\xx_{\bmm}$.
With $\gamma > 0$, we minimize
\begin{equation}
   \mathcal{L}_{\mathrm{DISCO}}^{\mathrm{mask}}(\btheta) := \mathcal{L}_{\mathrm{DISCO}}(\btheta) + \gamma L_{\mathrm{DISCO}}^{\mathrm{mask}}(\btheta)
\end{equation}
and with $\varphi := p(t) p(\mm)\, p_t(\xx_{t, \bmm}, \xx_{\mm})\,p_0(\xx_{\bmm} \cond \xx_{t,\bmm},\xx_{\mm})$ we define $L_{\mathrm{DISCO}}^{\mathrm{mask}}(\btheta)$ as
\begin{equation}
   \mathbb{E}_{\varphi}\!\Big[\lambda(t)\,\|\nabla_{\xx_{t, \bmm}}\log q_0(\xx_{t,\bmm}\mid \xx_{\bmm}) - P_{\bmm}s_\theta(\xx_{t, \bmm}, \xx_{\mm})\|^2_2\Big],
\end{equation}
where $P_{\bmm}$ projects onto the noised coordinates.
The global minimizer is $\nabla_{\xx} \log p_0(\xx)$ for every $\gamma \geq 0$; we use $\gamma = 0.5$ throughout our experiments.
Similar to \eqref{eq:posterior_empirical}, we draw from the conditioned ``zero-level'' posterior $p_0(\xx_{\bmm} \cond \xx_{t,\bmm},\xx_{\mm}) \propto q_0(\xx_{t,\bmm} \cond \xx_{\bmm}) p_{\text{emp}}(\xx_{\bmm} \cond \xx_{\mm})$. When $\mm \neq \mathbf{0}$, $p_{\text{emp}}(\xx_{\bmm} \cond \xx_{\mm})$ will typically place all its mass on the ``helper'' data point $\xx'_{\bmm}$ that was used to produce $\xx_{t,\bmm}$, making posterior sampling trivial: One just takes $\xx'_{\bmm}$ without materializing the posterior at all (see Appendix \ref{app:knn}).
\section{Related Work}
\label{sec:related_work}

\textbf{Time-Independence in Score-Based Models.} 
Most similar in spirit to DISCO is the work by Li et al. \citep{li2023learning}, who share the idea of only learning $\nabla_{\xx} \log p_0(\xx)$ using a score-matching objective. 
However, they do not minimize $\mathcal{L}_{\mathrm{DISCO}}$, but a variant which they term \emph{multiscale denoising score matching (MDSM)}, which is $\mathcal{L}_{\mathrm{DISCO}}$ when (incorrectly) setting $q(t, \xx_t, \xx) := p(t) \pd(\xx) p_t(\xx_t \cond \xx)$ in \eqref{eq:disco}.
This objective in fact learns $s_{\btheta}^*(\xx_t) = \mathbb{E}_{p(t \cond \xx_t)}\left[ \frac{\sigma(t)^2}{\sigma(0)^2} \nabla_{\xx_t} \log p_t(\xx_t) \right]$, i.e.~a \emph{reweighted posterior average over $p_t$ scores} (VE noise, $\lambda(t) = 1$); it differs from DISCO only in drawing $\xx$ before $\xx_t$. Thus, the claim of \citep{li2023learning} that $s_{\btheta}^*$ only learns the score of $p_0$ is erroneous (see Appendix \ref{proof:mdsm}); their motivation is also analyzing diffusion training rather than conditional sampling.
Other attempts to simplify or remove the time dependence in diffusion models---such as the scaled architecture of \citet{song2020improved} or the fully time-independent training of \citet{sun2025noise}---likewise do not recover $\nabla_{\xx} \log p_0(\xx)$, but instead yield posterior averages over noise-level scores (see Appendix \ref{app:sun}).

\textbf{Conditional Sampling in Diffusion Models.} Many approximations to the true conditional $p_0(\xx_0^u \cond \xx_0^{c})$ have been proposed: \citet[App.~I]{song2020sde} introduce the \emph{replacement method}, a popular heuristic that estimates $\nabla_{\xx_t^u} \log p_t(\xx_t^{u} \cond \xx_0^{c})$, i.e.~the conditional score at time $t$, as $\nabla_{\xx_t^u} \log p_t(\xx_t^{u} \cond \hat{\xx}_t^{c})$
where $\hat{\xx}_t^{c}$ is drawn from $p_t(\xx_t^c \cond \xx_0^c) = \mathcal{N}(\xx_t^c; \alpha(t)\xx_0^c, \sigma(t)^2I)$. This approximation enjoys no theoretical guarantees and often fails to produce samples coherent with the conditioning information \citep{ho2022video}. 

\emph{Gradient guidance} \citep{ho2022video} relies on the fact that $\nabla_{\xx_t} \log p_t(\xx_t \cond \xx_0^c) = \nabla_{\xx_t} \log p_t(\xx_0^c \cond \xx_t) + \nabla_{\xx_t} \log p_t(\xx_t)$. 
While $\nabla_{\xx_t} \log p_t(\xx_t)$ is known via $s_{\btheta}$, the intractable quantity $p_t(\xx_0^c \cond \xx_t)$ is approximated, often by $\mathcal{N}(\xx_0^c; \Omega(\hat{\xx}_{\btheta}(\xx_t, t)), \sigma(t)^2I)$, where $\hat{\xx}_{\btheta}(\xx, t) = \xx + \sigma(t)^2 s_{\btheta}(\xx, t)$ is the ``denoised'' input and $\Omega$ selects the observed coordinates. \emph{Diffusion posterior sampling} (DPS) \citep{chung2023diffusion} uses a closely related, normalized likelihood gradient; we compare to both.
Evaluating it requires backpropagating through the network, and the estimates can be unreliable \citep{zhang2023towards} and do not provide guarantees.

\citet{wu2023practical} introduced the \emph{twisted diffusion sampler} (TDS), which uses gradient guidance in a twisted sequential Monte Carlo (SMC) procedure as an approximation to the (unknown) optimal twisting function. 
Due to this, the sampler will not produce exact samples for any \emph{finite} number of simulated particles. In contrast, DISCO guarantees asymptotically exact samples from arbitray model conditionals, even when simulating \emph{a single particle}.

\textbf{Inverse Problems.}
PnP-DM \citep{wu2024principled}, DPG \citep{tang2024solving} and APMC \citep{sun2024provable} sample $p_{\btheta}(\xx \cond \mathbf{y}) \propto p(\mathbf{y} \cond \xx) \, p_{\btheta}(\xx)$ for a known, noisy forward model and a pretrained prior. They are complementary to DISCO, which addresses how to \emph{train} a joint model whose clean-coordinate conditionals are directly accessible (Appendix~\ref{app:inverse}). Our conditioning corresponds to the noise-free likelihood $\delta(\mathbf{y} - \Omega \xx)$, whose posterior is singular on $\mathbb{R}^D$; DISCO sidesteps this entirely by sampling only the unobserved coordinates.

\textbf{Directly Learning Conditionals.}
An alternative line of work \citep{shi2022conditional,kerrigan2024dynamic,baptista2024conditional} learns models that directly sample from conditional distributions, without parameterizing a coherent joint density. DISCO instead derives all conditionals from one joint model, which guarantees their consistency.

\section{Experiments}
\label{sec:experiments}

\subsection{Low-Dimensional Setting}
\label{sec:lowD}

We train both an energy-based diffusion model and an energy-based DISCO model on several 2D distributions (Moons, Checkerboard, and Rings) by minimizing $\mathcal{L}_{\mathrm{DM}}$ \eqref{eq:dm} and $\mathcal{L}_{\text{DISCO}}$ \eqref{eq:disco}, respectively. During training, we draw exact posterior samples from $p_0(\xx \cond \xx_t)$ and do not notice any significant slowdown compared to diffusion model training.
Appendix~\ref{app:minibatch_ablation} repeats this with the mini-batch posterior used in high dimensions (batches of $64$--$512$): model fit degrades mainly on Rings, while inference quality is unaffected.

\begin{figure*}[t]
\begin{minipage}[t]{0.49\textwidth}
    \centering
    \captionof{table}{\textbf{Quantitative Evaluation on 2D Distributions:} \emph{Model Fit} estimates the Wasserstein-1 distance $W_1\left(p_0(\xx), p_{\btheta}(\xx)\right)$ (diffusion: ancestral sampling; DISCO: tempered SMC). \emph{Inference Quality} estimates $\mathbb{E}_{p_0(x_1)} [ W_1(p_{\btheta}(x_2 \cond x_1), \hat{p}_{\btheta}(x_2 \cond x_1)) ]$, with $p_{\btheta}$ the exact \emph{model} conditional (rejection sampling) and $\hat{p}_{\btheta}$ the heuristic or, for DISCO, SMC on \eqref{eq:conditional_score}. Mean $\pm\, 2\sigma$ over $30$ independent runs (model fit) or $30$ conditioning values $x_1 \sim p_0(x_1)$ (inference quality).}
    \vspace{0.5em}
    \label{tab:w1_results_2d}
    \resizebox{\linewidth}{!}{
    \begin{tabular}{lccc}
    \toprule
     Method & Moons & Checkerboard & Rings \\
    \midrule
    \rowcolor[gray]{0.9} \multicolumn{4}{l}{\emph{Model Fit (joint)}} \\
     Diff. Model & \textbf{0.062 $\pm$ 0.009} & $0.057 \pm 0.004$ & $0.081 \pm 0.003$ \\
     DISCO & $0.068 \pm 0.022$ & \textbf{0.029 $\pm$ 0.008} & \textbf{0.066 $\pm$ 0.026} \\
    \midrule
    \rowcolor[gray]{0.9} \multicolumn{4}{l}{\emph{Inference Quality}} \\
     Replace & $0.363 \pm 0.409$ & $0.141 \pm 0.129$ & $0.099 \pm 0.074$ \\
     Grad. Guid. & $0.417 \pm 0.487$ & $0.165 \pm 0.129$ & $0.117 \pm 0.077$ \\
     TDS & $0.326 \pm 1.016$ & $0.120 \pm 0.165$ & $0.082 \pm 0.067$ \\
     DPS & $0.219 \pm 0.345$ & $0.123 \pm 0.103$ & $0.085 \pm 0.074$ \\
     DISCO & \textbf{0.024 $\pm$ 0.025} & \textbf{0.015 $\pm$ 0.012} & \textbf{0.042 $\pm$ 0.040} \\
    \bottomrule
    \end{tabular}}

\end{minipage}\hfill
\begin{minipage}[t]{0.49\textwidth}
\centering
\vspace{0pt}
\includegraphics[width=\linewidth]{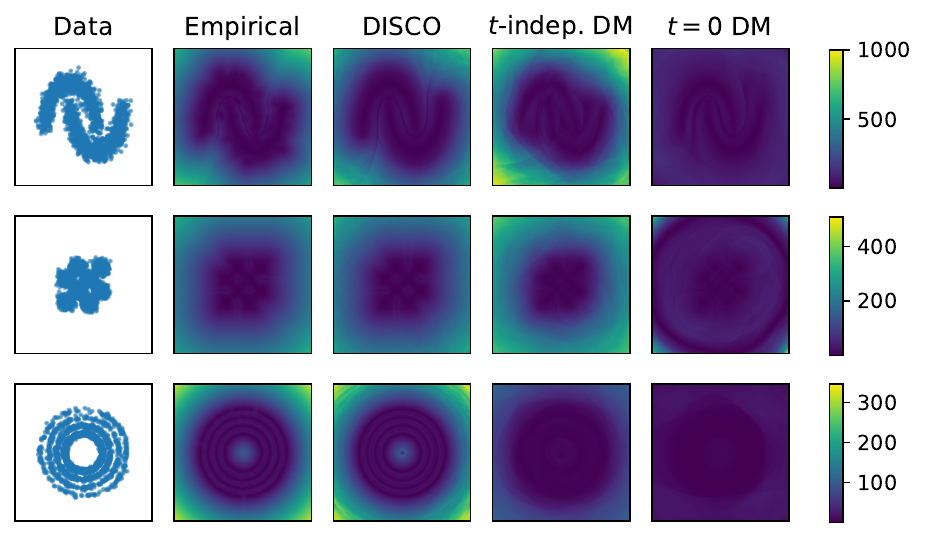}
\caption{\textbf{Qualitative Results on 2D Distributions:} $L_2$ norms of the scores of the empirical distribution, an energy-based DISCO model, an energy-based DM without access to $t$ (as in \citet{sun2025noise}), and an energy-based DM at $t = 0$; samples from $p_0$ on the left. Only DISCO recovers the true score magnitudes.}
\label{fig:2d_scores}
\end{minipage}
\end{figure*}

Figure~\ref{fig:2d_scores} compares the $L_2$ norms of the learned score fields. Only DISCO accurately recovers the scores of the true empirical distribution, while the diffusion model at $t=0$ heavily underestimates score magnitudes, especially far from the training data. This is expected: diffusion spreads the generative process over many time-dependent fields, and a time-independent diffusion model \citep{sun2025noise} learns an average of $p_t$ scores.

We then evaluate conditional sampling from \( p_{\boldsymbol{\theta}}(x_2 \cond x_1) \). As reference, we use exact conditional samples of each \emph{model}, obtained by rejection sampling from its joint samples; this isolates each model's consistency with its own learned joint.
Figure~\ref{fig:low_dim_inference} shows that only DISCO produces faithful conditional samples that preserve the relative weights of the Gaussian components, while all other methods fail. Note that the core issue is \emph{not} a bad fit to $\pd$: The diffusion model's \emph{true} conditionals fit the data well, but are computationally intractable to access. 

Table~\ref{tab:w1_results_2d} confirms this quantitatively: while DISCO performs on par with diffusion models in terms of model fit, it systematically outperforms all diffusion-based conditioning heuristics in inference quality, providing asymptotically---and practically---exact conditional samples.
Both model classes use comparable compute ($100$ ancestral steps vs.\ tempered SMC with HMC).
Further experimental details, including model architectures and 
sampler hyperparameters, are provided in Appendix~\ref
{app:experimental_details}.

\subsection{Image Datasets}

\begin{table*}[t]
\centering
\caption{
\textbf{Quantitative Inpainting Results:} On both CIFAR-10 and FFHQ-64, DISCO often outperforms post-hoc inpainting heuristics (Replacement, Gradient Guidance, DPS, RePaint and TDS) applied to a pre-trained diffusion model with EDM architecture. \emph{EDM Masked} (gray) learns separate conditional scores at each noise level and is a strong task-specific reference, but its conditionals need not be consistent with a single joint distribution. DISCO outperforms it on CIFAR-10 and comes close on FFHQ-64 while deriving all conditionals from a single score field (consistent with a joint in the limit of an exactly learned score), and has better unconditional samples on FFHQ-64 (see Table~\ref{tab:fid}).
Bold: best among methods that sample conditionals of a single joint model; \emph{EDM Masked} is excluded.
}
\scriptsize
\setlength{\tabcolsep}{4.8pt}
\renewcommand{\arraystretch}{1.10}
\begin{tabular}{lcccccccccccc}
\toprule
\textbf{CIFAR-10} & \multicolumn{2}{c}{\textit{Wide}} & \multicolumn{2}{c}{\textit{Narrow}} & \multicolumn{2}{c}{\textit{Super-Resolve 2x}} & \multicolumn{2}{c}{\textit{Altern. Lines}} & \multicolumn{2}{c}{\textit{Half}} & \multicolumn{2}{c}{\textit{Expand}} \\
Method & LPIPS$\downarrow$ & SSIM$\uparrow$ & LPIPS$\downarrow$ & SSIM$\uparrow$ & LPIPS$\downarrow$ & SSIM$\uparrow$ & LPIPS$\downarrow$ & SSIM$\uparrow$ & LPIPS$\downarrow$ & SSIM$\uparrow$ & LPIPS$\downarrow$ & SSIM$\uparrow$ \\
\midrule
Replacement & 0.255 & 0.658 & 0.064 & 0.916 & 0.431 & 0.516 & 0.187 & 0.767 & 0.296 & 0.549 & 0.607 & 0.094 \\
Grad. Guidance & 0.203 & 0.732 & 0.036 & 0.943 & 0.196 & 0.727 & 0.082 & 0.860 & 0.262 & 0.600 & 0.581 & 0.125 \\
DPS & 0.190 & 0.752 & 0.032 & 0.948 & 0.130 & 0.793 & 0.065 & 0.882 & 0.261 & 0.614 & 0.552 & 0.210 \\
RePaint & 0.225 & 0.732 & 0.078 & 0.914 & 0.339 & 0.676 & 0.184 & 0.806 & 0.285 & 0.593 & 0.627 & 0.140 \\
TDS & 0.232 & 0.760 & 0.053 & 0.937 & 0.255 & 0.746 & 0.124 & 0.857 & 0.308 & 0.623 & 0.630 & \textbf{0.230} \\
\midrule
DISCO & \textbf{0.161} & \textbf{0.790} & \textbf{0.018} & \textbf{0.971} & \textbf{0.120} & \textbf{0.831} & \textbf{0.053} & \textbf{0.899} & \textbf{0.227} & \textbf{0.630} & \textbf{0.535} & 0.170 \\
\midrule
\rowcolor[gray]{0.92} \emph{EDM Masked} & 0.165 & 0.765 & 0.025 & 0.954 & 0.113 & 0.789 & 0.060 & 0.881 & 0.232 & 0.624 & 0.539 & 0.170 \\
\midrule
\toprule
\textbf{FFHQ} & \multicolumn{2}{c}{\textit{Wide}} & \multicolumn{2}{c}{\textit{Narrow}} & \multicolumn{2}{c}{\textit{Super-Resolve 2x}} & \multicolumn{2}{c}{\textit{Altern. Lines}} & \multicolumn{2}{c}{\textit{Half}} & \multicolumn{2}{c}{\textit{Expand}} \\
Method & LPIPS$\downarrow$ & SSIM$\uparrow$ & LPIPS$\downarrow$ & SSIM$\uparrow$ & LPIPS$\downarrow$ & SSIM$\uparrow$ & LPIPS$\downarrow$ & SSIM$\uparrow$ & LPIPS$\downarrow$ & SSIM$\uparrow$ & LPIPS$\downarrow$ & SSIM$\uparrow$ \\
\midrule
Replacement & 0.192 & 0.673 & 0.048 & 0.916 & 0.410 & 0.562 & 0.106 & 0.851 & 0.218 & 0.595 & 0.502 & 0.150 \\
Grad. Guidance & 0.140 & 0.749 & 0.027 & 0.941 & 0.114 & 0.797 & 0.037 & 0.914 & 0.178 & 0.665 & 0.481 & 0.186 \\
DPS & 0.123 & 0.774 & \textbf{0.023} & 0.949 & \textbf{0.059} & \textbf{0.859} & 0.026 & \textbf{0.932} & 0.168 & 0.682 & \textbf{0.429} & \textbf{0.269} \\
RePaint & 0.136 & 0.767 & 0.027 & 0.942 & 0.107 & 0.820 & 0.057 & 0.903 & 0.184 & 0.670 & 0.491 & 0.211 \\
TDS & 0.142 & \textbf{0.786} & 0.028 & \textbf{0.951} & 0.136 & 0.829 & 0.059 & 0.916 & 0.189 & \textbf{0.697} & 0.488 & 0.209 \\
\midrule
DISCO & \textbf{0.112} & 0.784 & 0.029 & 0.945 & 0.067 & 0.842 & \textbf{0.024} & 0.931 & \textbf{0.161} & 0.686 & 0.431 & 0.250 \\
\midrule
\rowcolor[gray]{0.92} \emph{EDM Masked} & 0.108 & 0.787 & 0.018 & 0.956 & 0.050 & 0.865 & 0.022 & 0.935 & 0.164 & 0.680 & 0.435 & 0.247 \\
\bottomrule
\end{tabular}
\label{tab:inpainting-results}
\end{table*}
\begin{table}[ht]
    \centering
    \caption{FID scores on \emph{unconditional} image generation.} 
    \label{tab:fid}
    \setlength{\tabcolsep}{8pt} 
    \begin{tabular}{lccc}
        \toprule
        Dataset & EDM & EDM Masked & DISCO \\
        \midrule
        CIFAR-10 & 1.97 & 2.85 & 3.18 \\
        FFHQ-64  & 2.39 & 3.18 & 2.54 \\
        \bottomrule
    \end{tabular}
\end{table}

\begin{figure*}
    \centering
    \includegraphics[width=1.0\linewidth]{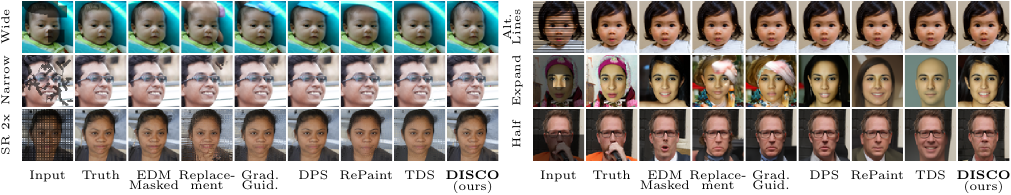}
    \caption{\textbf{Qualitative Inpainting Results:} Comparison of inpainting methods on FFHQ-64 with the samplers of Table~\ref{tab:inpainting-results}; \emph{Input} dims the unobserved pixels, \emph{Truth} shows the original image. Best viewed zoomed in. More results (conditional and unconditional) can be found in Appendix \ref{app:high_dim_qualitative}.}
    \label{fig:ffhq_inpainting}
\end{figure*}

In this section, our aim is \emph{not} state-of-the-art image generation, but rather to show that DISCO (i) is competitive with SOTA DM-based generative models in terms of model fit, while (ii) \emph{showing substantial advantages in terms of faithful conditioning tasks.}
To this end, we train unconstrained score models (i.e., not derived from an energy-based model) with DISCO on CIFAR-10 \citep{krizhevsky2009learning} and FFHQ-64 \citep{ffhq}.
We build on the popular \emph{EDM} framework \citep{edm}, adapting its architecture to be time-independent and training it by minimizing $\mathcal{L}_{\mathrm{DISCO}}^{\mathrm{mask}}$ with $\gamma = 0.5$ and the mini-batch approximation to sample from the posterior $p_0(\xx \cond \xx_t)$ (see Appendix~\ref{app:edm_disco} for details).
Training hyperparameters are inherited from EDM; DISCO was about as sensitive to $p(t)$ and $\lambda(t)$ as diffusion training, and robust to the mask distribution.

As a baseline, we train \emph{EDM Masked}, which is equivalent to the original diffusion model with \emph{EDM} architecture \cite{edm}, except that it minimizes $\mathcal{L}_{\mathrm{DM}}^{\mathrm{mask}}$ (defined in Appendix \ref{app:masked_dm}) with $\gamma = 0.5$ instead of $\mathcal{L}_{\mathrm{DM}}$ \eqref{eq:dm}. 
$\mathcal{L}_{\mathrm{DM}}^{\mathrm{mask}}$ directly fits conditional scores at every noise level, with no guarantee that they are \emph{consistent} with a single joint distribution; EDM Masked is thus a heuristic rather than a principled probabilistic model (Appendix~\ref{app:masked_dm}).
Inconsistent conditionals can still inpaint well: EDM Masked is a strong \emph{task-specific} model, whereas DISCO yields one coherent joint, which matters when many queries must be answered consistently, e.g., in uncertainty quantification or scientific modeling.
As inpainting is just one application of conditioning, we compare to the same general-purpose heuristics as before \citep{song2020sde,ho2022video,kawar2022denoising}, plus RePaint \citep{repaint} and DPS \citep{chung2023diffusion} (Appendix~\ref{app:experimental_details}).

To sample unconditionally and conditionally from DISCO or \emph{EDM Masked}, we use the second-order Heun sampler \citep{edm}; DISCO's settings were tuned, while EDM Masked uses its default settings except for using DISCO's number of steps for FFHQ-64 inpainting (Appendix~\ref{app:experimental_details}). 
Table~\ref{tab:fid} reports unconditional (joint) sample quality: While DISCO shows a gap to diffusion models on CIFAR-10, it---perhaps surprisingly---outperforms the \emph{EDM Masked} baseline on FFHQ-64, demonstrating that learning a \emph{single} score vector field can yield high visual sample quality compared to state-of-the-art diffusion-based training.

Inpainting results are reported qualitatively in Figure~\ref{fig:ffhq_inpainting} and quantitatively in Table~\ref{tab:inpainting-results}, using LPIPS \citep{lpips} and SSIM \citep{ssim}. DISCO outperforms almost all post-hoc heuristics on CIFAR-10, is on par with the strongest one, DPS, on FFHQ-64, and outperforms \emph{EDM Masked} on CIFAR-10 while coming close on FFHQ-64, with all conditionals derived from a single score field (consistent with a joint in the limit of an exactly learned score).
Remaining differences (e.g., on \emph{Super-Resolve 2x}) may stem from the mini-batch posterior and untuned training hyperparameters \citep{sun2025noise}.

\begin{figure}[t]
\centering
\includegraphics[width=\linewidth]{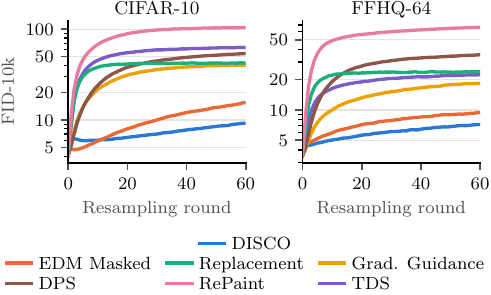}
\caption{\textbf{Resampling Consistency Test:} FID-10k (log scale) under repeated resampling of random masks, starting from each model's samples.}
\label{fig:gibbs}
\end{figure}

\paragraph{Consistency of Conditionals.}
To test \emph{consistency}, we start from $10{,}000$ samples of each model and repeatedly resample the pixels outside a random training mask with the method's conditional sampler. For conditionals of a single joint, this blocked Gibbs sampler leaves the sample distribution, and hence FID, invariant. All post-hoc samplers drift by a factor of 5 or more, \emph{EDM Masked} by 3.2 (CIFAR-10: $4.9 \to 15.5$ after 60 rounds), and DISCO least ($6.6 \to 9.2$; FFHQ-64: $4.2 \to 7.2$; Figure~\ref{fig:gibbs}). Without Metropolis correction, the drift mixes model and sampler error; see  Appendix~\ref{app:gibbs} for details and qualitative results.

\FloatBarrier
\section{Conclusion}
\label{sec:conclusion}

In this paper, we challenge the prevailing belief that diffusion processes are essential for training effective score-based generative models. 
We introduce DISCO, a diffusion-free score matching framework that avoids time-indexed score fields in favor of learning a single, time-independent score function. 
It is close to diffusion models in sample quality and, more importantly, provides a principled foundation for asymptotically exact conditional sampling.

\subsection*{AI Use Statement}

In this work, we used generative AI tools to assist in implementing methods, namely the code for our experiments and for the baseline methods, and in running these experiments. All LLM-generated code was verified for correctness by the authors.
We did not use generative AI tools to generate research ideas or hypotheses, to develop theoretical models, to formulate mathematical claims, or to provide ingredients for or write proofs; generating synthetic data sets, translation, cleaning or reformatting data sets, and qualitative data analysis are not applicable to this work.
We also used generative AI tools to suggest and give feedback on experiments, to suggest experimental parameters, and to assist in interpreting experimental results, all of which were reviewed by the authors.
Additionally, we used generative AI tools to assist in writing and editing the paper, in creating figures, and in formatting references.
We have reviewed all AI-assisted work, and we take full responsibility for the final content of this work, including text, claims, or artifacts produced with the aid of generative AI.

\subsubsection*{Acknowledgements}
This research was funded in whole or in part by the Austrian Science Fund (FWF) 10.55776/COE12. This project has received funding from the European Union's EIC Pathfinder Challenges 2022 programme under grant agreement No 101115317 (NEO).

\bibliography{disco}

\clearpage
\appendix
\thispagestyle{empty}

\onecolumn
\aistatstitle{Diffusion-free Score Matching for Exact Conditional Sampling: \\
Supplementary Materials}

\section{Masked Diffusion Model Training} 
\label{app:masked_dm}
To provide a baseline for the masked DISCO training, we analogously define a masked diffusion model training objective. With
\begin{equation*}
\varphi(t, \mm, \xx, \xx_{t,\bmm}) := \\
p(t)\,p(\mm)\,\pd(\xx)\,q_t(\xx_{t,\bmm} \mid \xx_{\bmm})
\end{equation*}
and some $\gamma > 0$, we again minimize
\begin{equation}
   \mathcal{L}_{\mathrm{DM}}^{\mathrm{mask}}(\btheta) := \mathcal{L}_{\mathrm{DM}}(\btheta) + \gamma L_{\mathrm{DM}}^{\mathrm{mask}}(\btheta)
\end{equation}
where $L_{\mathrm{DM}}^{\mathrm{mask}}(\btheta)$ is defined as 
\begin{equation}
\mathbb{E}_{\varphi}\!\left[
\left\| \nabla_{\xx_{t,\bmm}} \log q_t(\xx_{t,\bmm} \mid \xx_{\bmm}) - P_{\bmm} s_{\btheta}(\xx_{t,\bmm}, \xx_{\mm}, t, \mm) \right\|^2_2 \right].
\end{equation}

For the empty mask $\mm = \mathbf{0}$, the global optimum of $\mathcal{L}_{\mathrm{DM}}^{\mathrm{mask}}$ is $s_{\btheta}^*(\xx_t, t, \mm = \mathbf{0}) = \nabla_{\xx_t} \log p_t(\xx_t)$. For other $\mm$, the global optimum is $P_{\bmm} s_{\btheta}^*(\xx_t, t, \mm) = \nabla_{\xx_{t, \bmm}} \log p_t(\xx_{t, \bmm} \mid \xx_{\mm})$.
Note that this is merely circumventing the problem of probabilistic inference by directly trying to learn the otherwise intractable quantities. Therefore, we have no guarantee that the conditionals we learn are \emph{consistent}, i.e., that there exists a joint distribution whose conditionals coincide with the learned ones. Clearly, this breaks sound probabilistic reasoning, and hence, these models should also be considered a heuristic baseline, rather than a principled approach.

\section{Proofs}
\subsection{DISCO Objective}
\label{app:disco_loss_proof}
Let $p(t)$ be a prior distribution over a ``time'' parameter\footnote{We want to stress that it has only the meaning of time in diffusion models, while in DISCO it indexes a family of successively noisier proposal distributions.} $t \in [0, T]$, let $\pd$ denote the data distribution, and let $\alpha: [0, T] \to \mathbb{R}_{>0}$ and $\sigma: [0, T] \to \mathbb{R}_{>0}$ be positive functions of time. 
Given the distributions $p(t, \xx, \xx_t) := p(t)\pd(\xx)q_t(\xx_t \cond \xx)$ with $q_t(\xx_t \cond \xx) := \mathcal{N}(\xx_t; \alpha(t)\xx, \sigma(t)^2I)$ and $q(t, \xx, \xx_t) := p(t)p_t(\xx_t)p_0(\xx \cond \xx_t)$ with $p_t(\xx_t) = \int q_t(\xx_t \cond \xx)\pd(\xx) \,\mathrm{d}\xx$ and 
\[ p_0(\xx \cond \xx_t) = \frac{q_0(\xx_t \cond \xx)\pd(\xx)}{p_0(\xx_t)}, \]
we will show that the \emph{DISCO Loss}
\begin{equation}
\mathcal{L}_{\mathrm{DISCO}}(\btheta) := \mathbb{E}_{q(t,\xx, \xx_t)} \left[ \lambda(t) \|\nabla_{\xx_t} \log q_0(\xx_t \cond \xx) - s_{\btheta}(\xx_t) \|^2_2 \right]
\end{equation}
is equivalent to 
\begin{equation}
\mathcal{F}_{\mathrm{mix}}(p_0 \, \| \, s_{\btheta}) = \mathbb{E}_{p(t)}\left[ \lambda(t) \mathcal{F}_{p_t}(p_0 \, \| \, s_{\btheta}) \right] = \mathbb{E}_{p(t)}\mathbb{E}_{p_t(\xx_t)} \left[ \lambda(t) \|\nabla_{\xx_t} \log p_0(\xx_t) - s_{\btheta}(\xx_t) \|^2_2 \right]
\end{equation}
up to an additive constant independent of $\btheta$, where $p_0$ denotes the slightly Gaussian-perturbed data distribution as defined in the main text.

\begin{proof}
We see that 
\begin{align*}
\mathcal{L}_{\mathrm{DISCO}}(\btheta) &= \mathbb{E}_{q(t,\xx, \xx_t)} \left[ \lambda(t) \|\nabla_{\xx_t} \log q_0(\xx_t \cond \xx) - s_{\btheta}(\xx_t) \|^2_2 \right] \\
&= \mathbb{E}_{p(t)p_t(\xx_t)} \left[ \lambda(t) \, \mathbb{E}_{p_0(\xx \cond \xx_t)} \left[ \|\nabla_{\xx_t} \log q_0(\xx_t \cond \xx) - s_{\btheta}(\xx_t) \|^2_2 \right] \right]
\end{align*}
We have
\begin{align*}
&\mathbb{E}_{p_0(\xx \cond \xx_t)} \left[ \|\nabla_{\xx_t} \log q_0(\xx_t \cond \xx) - s_{\btheta}(\xx_t) \|^2_2 \right] \\
= \ &\mathbb{E}_{p_0(\xx \cond \xx_t)} \left[ \|\nabla_{\xx_t} \log q_0(\xx_t \cond \xx) \|^2_2 - 2 \nabla_{\xx_t} \log q_0(\xx_t \cond \xx)^\top s_{\btheta}(\xx_t) + \|s_{\btheta}(\xx_t)\|^2_2 \right]\\
= \ &c_1 - 2 \mathbb{E}_{p_0(\xx \cond \xx_t)} \left[\nabla_{\xx_t} \log q_0(\xx_t \cond \xx)\right]^\top s_{\btheta}(\xx_t) + \|s_{\btheta}(\xx_t)\|^2_2\\
= \ & c_2 + \| \mathbb{E}_{p_0(\xx \cond \xx_t)} \left[\nabla_{\xx_t} \log q_0(\xx_t \cond \xx) \right] - s_{\btheta}(\xx_t) \|^2_2 
\end{align*}
where $c_1, c_2$ are constants w.r.t. $\btheta$. We notice that
\begin{align*}
\mathbb{E}_{p_0(\xx \cond \xx_t)} \left[ \nabla_{\xx_t} \log q_0(\xx_t \cond \xx) \right] &= \int p_0(\xx \cond \xx_t) \nabla_{\xx_t} \log q_0(\xx_t \cond \xx) \,\mathrm{d}\xx \\
&= \int p_0(\xx \cond \xx_t) \frac{\nabla_{\xx_t} q_0(\xx_t \cond \xx)}{q_0(\xx_t \cond \xx) }  \,\mathrm{d}\xx \\
&= \int \frac{q_0(\xx_t \cond \xx)\pd(\xx)}{p_0(\xx_t)} \frac{\nabla_{\xx_t} q_0(\xx_t \cond \xx)}{q_0(\xx_t \cond \xx) }  \,\mathrm{d}\xx \\
&= \int \frac{\pd(\xx) \nabla_{\xx_t} q_0(\xx_t \cond \xx)}{p_0(\xx_t)} \,\mathrm{d}\xx \\
&= \frac{1}{p_0(\xx_t)} \int \pd(\xx) \nabla_{\xx_t} q_0(\xx_t \cond \xx) \,\mathrm{d}\xx \\
&= \frac{1}{p_0(\xx_t)} \nabla_{\xx_t} \int \pd(\xx) q_0(\xx_t \cond \xx) \,\mathrm{d}\xx \\
&= \frac{1}{p_0(\xx_t)} \nabla_{\xx_t} p_0(\xx_t) \\
&= \nabla_{\xx_t} \log p_0(\xx_t)
\end{align*}
and hence, 
\begin{align*}
\| \mathbb{E}_{p_0(\xx \cond \xx_t)} \left[\nabla_{\xx_t} \log q_0(\xx_t \cond \xx) \right] - s_{\btheta}(\xx_t) \|^2_2 = \| \nabla_{\xx_t} \log p_0(\xx_t) - s_{\btheta}(\xx_t) \|^2_2
\end{align*}
which implies that
\begin{align*}
\mathcal{L}_{\mathrm{DISCO}}(\btheta) &= \mathbb{E}_{p(t)p_t(\xx_t)} \left[ \lambda(t) \, \mathbb{E}_{p_0(\xx \cond \xx_t)} \left[ \|\nabla_{\xx_t} \log q_0(\xx_t \cond \xx) - s_{\btheta}(\xx_t) \|^2_2 \right] \right] + \mathrm{const.}\\
&= \mathbb{E}_{p(t)p_t(\xx_t)} \left[ \lambda(t) \, \| \nabla_{\xx_t} \log p_0(\xx_t) - s_{\btheta}(\xx_t) \|^2_2 \right] + \mathrm{const.} \\
&= \mathbb{E}_{p(t)}\left[ \lambda(t) \mathcal{F}_{p_t}(p_0 \, \| \, s_{\btheta}) \right]  + \mathrm{const.} 
\end{align*}
which concludes the proof.
\end{proof}

\subsection{Multiscale Denoising Score Matching}
\label{app:mdsm}

We show that the \emph{multiscale denoising score matching (MDSM)} \citep{li2023learning} objective 
\begin{align}
\mathcal{L}_{\mathrm{MDSM}}(\btheta) &= \mathbb{E}_{p(t)\pd(\xx)q_t(\xx_t \cond \xx)} \left[ \lambda(t) \, \|\nabla_{\xx_t} \log q_0(\xx_t \cond \xx) - s_{\btheta}(\xx_t) \|^2_2 \right]
\end{align}
has the minimizer $s^*_{\btheta}(\xx_t) = \mathbb{E}_{p(t \cond \xx_t)}\left[ \frac{\sigma(t)^2}{\sigma(0)^2} \nabla_{\xx_t} \log p_t(\xx_t) \right]$ when $\lambda(t) = 1$ and $\alpha(t) = 1$ for all $t$ (variance exploding).

\begin{proof}
\label{proof:mdsm}
For convenience, we assume $\lambda(t) = 1$, as this can always be subsumed into the prior $p(t)$ without affecting the minimizer. Moreover, we assume $\alpha(t) = 1$.
With $p(t,\xx,\xx_t) = p(t)\pd(\xx)q_t(\xx_t \cond \xx)$, we denote with $p(\xx_t)$ the marginal over $\xx_t$ (not to be confused with $p_t(\xx_t)$, which conditions on $t$). We have
\begin{align}
\mathcal{L}_{\mathrm{MDSM}}(\btheta) &= \mathbb{E}_{p(t)\pd(\xx)q_t(\xx_t \cond \xx)} \left[ \|\nabla_{\xx_t} \log q_0(\xx_t \cond \xx) - s_{\btheta}(\xx_t) \|^2_2 \right] \\
&= \mathbb{E}_{p(\xx_t)p(t \cond \xx_t)p_t(\xx \cond \xx_t)} \left[ \|\nabla_{\xx_t} \log q_0(\xx_t \cond \xx) - s_{\btheta}(\xx_t) \|^2_2 \right] \\
&= \mathbb{E}_{p(\xx_t)p(t \cond \xx_t)} \left[ \|\mathbb{E}_{p_t(\xx \cond \xx_t)} \left[ \nabla_{\xx_t} \log q_0(\xx_t \cond \xx) \right] - s_{\btheta}(\xx_t) \|^2_2 \right] + \mathrm{const.} 
\end{align}
where the last step follows the same argument as in Section \ref{app:disco_loss_proof}.
With $R(\xx_t, t) := \mathbb{E}_{p_t(\xx \cond \xx_t)} \left[ \nabla_{\xx_t} \log q_0(\xx_t \cond \xx) \right]$ and repeating this argument, we see that
\begin{align}
\mathbb{E}_{p(\xx_t)p(t \cond \xx_t)}\left[ \| R(\xx_t, t)  - s_{\btheta}(\xx_t) \|^2_2 \right] = 
\mathbb{E}_{p(\xx_t)} \left[ \| \mathbb{E}_{p(t \cond \xx_t)} \left[ R(\xx_t, t) \right]  - s_{\btheta}(\xx_t) \|^2_2 \right] + \mathrm{const.}
\end{align}
where clearly, the minimizer is 
\begin{align}
s^*_{\btheta}(\xx_t) &= \mathbb{E}_{p(t \cond \xx_t)} \left[ R(\xx_t, t) \right] \\
&= \mathbb{E}_{p(t \cond \xx_t)p_t(\xx \cond \xx_t)} \left[ \nabla_{\xx_t} \log q_0(\xx_t \cond \xx) \right].
\end{align}
Expanding $\nabla_{\xx_t} \log q_0(\xx_t \cond \xx) = (\xx - \xx_t)/\sigma(0)^2$, we get
\begin{align}
s^*_{\btheta}(\xx_t) &= \mathbb{E}_{p(t \cond \xx_t)p_t(\xx \cond \xx_t)} \left[ \frac{\xx - \xx_t}{\sigma(0)^2}\right] = \mathbb{E}_{p(t \cond \xx_t)} \left[ \frac{\mathbb{E}_{p_t(\xx \cond \xx_t)}\left[\xx\right] - \xx_t}{\sigma(0)^2}\right] 
\end{align}
Via Tweedie's formula, we can express the posterior mean as $\mathbb{E}_{p_t(\xx \cond \xx_t)}\left[\xx\right] = \xx_t + \sigma(t)^2 \nabla_{\xx_t} \log p_t(\xx_t)$, and thus,
\begin{align}
s^*_{\btheta}(\xx_t) &= \mathbb{E}_{p(t \cond \xx_t)} \left[ \frac{\xx_t + \sigma(t)^2 \nabla_{\xx_t} \log p_t(\xx_t) - \xx_t}{\sigma(0)^2}\right] = \mathbb{E}_{p(t \cond \xx_t)} \left[ \frac{\sigma(t)^2}{\sigma(0)^2} \nabla_{\xx_t} \log p_t(\xx_t)\right] 
\end{align}
which concludes the proof.
\end{proof}
This shows that the claim made in \citet{li2023learning} that $s^*_{\btheta}(\xx)$ only learns $\nabla_{\xx}\log p_0(\xx)$ is incorrect.

\subsection{Masked DISCO Training}
\label{app:masked_disco_proof}

We prove that the global minimum of $\mathcal{L}_{\mathrm{DISCO}}^{\mathrm{mask}}(\btheta)$ is the same as that of $\mathcal{L}_{\mathrm{DISCO}}(\btheta)$.

\begin{proof}
Recall that 
\begin{align}
\mathcal{L}_{\mathrm{DISCO}}^{\mathrm{mask}}(\btheta) &= \mathcal{L}_{\mathrm{DISCO}}(\btheta) + \gamma L_{\mathrm{mask}}^{\mathrm{DISCO}}(\btheta)
\end{align}
where
\begin{align}
L_{\mathrm{mask}}^{\mathrm{DISCO}}(\btheta) &= \mathbb{E}_{\varphi(t, \mm, \xx_{t,\bmm},\xx_{\mm}, \xx_{\bmm})}\Big[\lambda(t)\,\|\nabla_{\xx_{t, \bmm}}\log q_0(\xx_{t,\bmm}\mid \xx_{\bmm}) - P_{\bmm}s_\theta(\xx_{t, \bmm}, \xx_{\mm})\|^2_2\Big]
\end{align}
and $\varphi(t, \mm, \xx_{t,\bmm},\xx_{\mm}, \xx_{\bmm}) = p(t)p(\mm)p_t(\xx_{t, \bmm}, \xx_{\mm})p_0(\xx_{\bmm} \cond \xx_{t,\bmm},\xx_{\mm})$.

Let $s_{\btheta}^*$ denote the global minimizer of $\mathcal{L}_{\mathrm{DISCO}}(\btheta)$. From Proof \ref{app:disco_loss_proof}, we know that 
\begin{equation}
s_{\btheta}^*(\xx_t) = \nabla_{\xx_t} \log p_0(\xx_t)
\end{equation}
We now show that $s_{\btheta}^*$ also minimizes $L_{\mathrm{mask}}^{\mathrm{DISCO}}(\btheta)$, for any $\gamma \geq 0$.
Analogously to Proof \ref{app:disco_loss_proof}, we have that
\begin{align*}
L_{\mathrm{mask}}^{\mathrm{DISCO}}(\btheta) &= \mathbb{E}_{\varphi(t, \mm, \xx_{t,\bmm},\xx_{\mm}, \xx_{\bmm})}\Big[\lambda(t)\,\|\nabla_{\xx_{t, \bmm}}\log q_0(\xx_{t,\bmm}\mid \xx_{\bmm}) - P_{\bmm}s_\theta(\xx_{t, \bmm}, \xx_{\mm})\|^2_2\Big] \\
&= \mathbb{E}_{p(t)p(\mm)p_t(\xx_{t,\bmm}, \xx_{\mm})}\Big[\lambda(t)\,\|\mathbb{E}_{p_0(\xx_{\bmm} \cond \xx_{t,\bmm},\xx_{\mm})}\left[\nabla_{\xx_{t, \bmm}}\log q_0(\xx_{t,\bmm}\mid \xx_{\bmm})\right] - P_{\bmm}s_\theta(\xx_{t, \bmm}, \xx_{\mm})\|^2_2\Big] + \mathrm{const.} 
\end{align*}
Note that the perturbation only acts on the coordinates where $\bmm$ is $1$ (i.e., the masked coordinates), and hence,
\begin{equation}
p_0(\xx_{t,\bmm}, \xx_{\bmm}, \xx_{\mm}) = \pd(\xx_{\bmm}, \xx_{\mm})\, q_0(\xx_{t,\bmm} \mid \xx_{\bmm}).
\end{equation}
Using this fact, we have
$$\begin{aligned}
&\mathbb{E}_{p_0(\xx_{\bmm} \mid \xx_{t,\bmm}, \xx_{\mm})}
\left[\nabla_{\xx_{t,\bmm}} \log q_0(\xx_{t,\bmm} \mid \xx_{\bmm})\right] \\
&\qquad= \int p_0(\xx_{\bmm} \mid \xx_{t,\bmm}, \xx_{\mm})
\,\nabla_{\xx_{t,\bmm}} \log q_0(\xx_{t,\bmm} \mid \xx_{\bmm})
\,\mathrm{d}\xx_{\bmm} \\[2pt]
&\qquad= \int 
\frac{p_0(\xx_{t,\bmm}, \xx_{\bmm}, \xx_{\mm})}
     {p_0(\xx_{t,\bmm}, \xx_{\mm})}
\cdot
\frac{\nabla_{\xx_{t,\bmm}} q_0(\xx_{t,\bmm} \mid \xx_{\bmm})}
     {q_0(\xx_{t,\bmm} \mid \xx_{\bmm})}
\,\mathrm{d}\xx_{\bmm} \\[2pt]
&\qquad= \frac{1}{p_0(\xx_{t,\bmm}, \xx_{\mm})}
\int \pd(\xx_{\bmm}, \xx_{\mm})
\,\nabla_{\xx_{t,\bmm}} q_0(\xx_{t,\bmm} \mid \xx_{\bmm})
\,\mathrm{d}\xx_{\bmm} \\[2pt]
&\qquad= \frac{1}{p_0(\xx_{t,\bmm}, \xx_{\mm})}
\,\nabla_{\xx_{t,\bmm}}
\int \pd(\xx_{\bmm}, \xx_{\mm})
\,q_0(\xx_{t,\bmm} \mid \xx_{\bmm})
\,\mathrm{d}\xx_{\bmm} \\[2pt]
&\qquad= \frac{1}{p_0(\xx_{t,\bmm}, \xx_{\mm})}
\,\nabla_{\xx_{t,\bmm}} p_0(\xx_{t,\bmm}, \xx_{\mm})
= \nabla_{\xx_{t,\bmm}} \log p_0(\xx_{t,\bmm} \mid \xx_{\mm})
\end{aligned}$$
which shows that the projection $P_{\bmm}s_{\btheta}(\xx_{t,\bmm}, \xx_{\mm})$ regresses the conditional score $\nabla_{\xx_{t,\bmm}} \log p_0(\xx_{t,\bmm} \mid \xx_{\mm})$.

Since $s_{\btheta}^*$ obeys
\begin{equation}
P_{\bmm}s_{\btheta}^*(\xx_{t,\bmm}, \xx_{\mm}) = \nabla_{\xx_{t,\bmm}} \log p_0(\xx_{t,\bmm}, \xx_{\mm}) = \nabla_{\xx_{t,\bmm}} \log p_0(\xx_{t,\bmm} \mid \xx_{\mm})
\end{equation}
for any mask $\mm$, this implies that $s_{\btheta}^*$ also minimizes $L_{\mathrm{mask}}^{\mathrm{DISCO}}(\btheta)$.
\end{proof}

\subsection{Time-Independent Diffusion Models}
\label{app:sun}

\paragraph{Scaled architecture.} \citet{song2020improved} proposed the parametrization $s_{\btheta}(\xx, t) := \varepsilon_{\btheta}(\xx) / \sigma(t)$, where $\varepsilon_{\btheta}$ is time-independent. This assumes that scores at different noise levels are simply scaled versions of one another, which is generally false: there exists no constant $c$ such that $\nabla_{\xx} \log p_{t_1}(\xx) = c \cdot \nabla_{\xx} \log p_{t_2}(\xx)$ for all $\xx$ and $t_1 \neq t_2$, except in the trivial case where $p_0$ is Gaussian. Hence this parametrization cannot recover the correct scores even with infinite capacity.

\paragraph{Fully time-independent diffusion training.} We show that minimizing $\mathcal{L}_{\mathrm{DM}}$ with a \emph{time-independent} score model $s_{\btheta}(\xx_t)$, i.e.,
\begin{align}
\mathcal{L}_{\mathrm{DM}}(\btheta) = \mathbb{E}_{t,\xx_0, \xx_t} \left[ \lambda(t) \, \|\nabla_{\xx_t} \log q_t(\xx_t \cond \xx_0) - s_{\btheta}(\xx_t) \|^2_2 \right],
\end{align}
leads to a minimizer $s^*_{\btheta}(\xx_t) = \mathbb{E}_{p(t \cond \xx_t)}\left[ \nabla_{\xx_t} \log p_t(\xx_t) \right]$ when $\lambda(t) = 1$ and $\alpha(t) = 1$ for all $t$.

\begin{proof}
As the proof looks almost identical to Proof \ref{proof:mdsm}, we will only briefly sketch it and refer the reader to \citet{sun2025noise} for more details.
With $R(\xx_t, t) := \mathbb{E}_{p_t(\xx \cond \xx_t)} \left[ \nabla_{\xx_t} \log q_t(\xx_t \cond \xx) \right]$, we again have that
\begin{align}
s^*_{\btheta}(\xx_t) &= \mathbb{E}_{p(t \cond \xx_t)} \left[ R(\xx_t, t) \right] = \mathbb{E}_{p(t \cond \xx_t)p_t(\xx \cond \xx_t)} \left[ \frac{\xx - \xx_t}{\sigma(t)^2}\right] 
\end{align}
Again via Tweedie's formula, we obtain
\begin{align}
s^*_{\btheta}(\xx_t) &= \mathbb{E}_{p(t \cond \xx_t)} \left[ \frac{\xx_t + \sigma(t)^2 \nabla_{\xx_t} \log p_t(\xx_t) - \xx_t}{\sigma(t)^2}\right] = \mathbb{E}_{p(t \cond \xx_t)} \left[\nabla_{\xx_t} \log p_t(\xx_t)\right] 
\end{align}
which concludes the proof.
\end{proof}

\pagebreak
\section{Posterior Sampling}
\label{app:knn}
When optimizing $\mathcal{L}_{\mathrm{DISCO}}$, we need to draw samples from the $t=0$ posterior 
\[ p_0(\xx \cond \xx_t) = \frac{q_0(\xx_t \cond \xx) \,\pd(\xx)}{p_0(\xx_t)}. \]
When we set $\pd(\xx) = \pe(\xx)$, we can draw \emph{exact} samples from $p_0(\xx \cond \xx_t)$: Given $\xx_t \in \mathbb{R}^D$, we compute $q(\xx_t \cond \xx^{(i)})$ for each $\xx^{(i)} \in \dat$, and sample $\xx$ from the normalized mass function over elements in $\dat$. Intuitively, since the perturbation kernel $q(\xx_t \cond \xx^{(i)})$ is an isotropic Gaussian, it will assign more probability mass to points $\xx^{(i)}$ that are close to $\xx_t$. 
This is distinct but reminiscent of the popular (minibatch) optimal transport techniques in the flow matching literature \citep{tong2023improving}.

Sampling from the posterior in this way needs $O(ND)$ operations, where $N = |\dat|$. In our low-dimensional experiments ($N=100,000$ and $D=2$) we draw exact posterior samples and do not observe any significant slowdown during model training. In our high-dimensional experiments (CIFAR-10, FFHQ-64), we draw approximate posterior samples by computing the mass function over a minibatch instead of the entire dataset.

Future work may explore utilizing techniques like \emph{Locality Sensitive Hashing} \citep{gionis1999similarity} or $k$-d Trees to efficiently get the $k$ nearest neighbors of $\xx_t$, and then compute the mass function over just these neighbors. If $\sigma(0)$ is sufficiently small, this will be a good approximation to the true posterior mass function over all elements in $\dat$.

When optimizing $\mathcal{L}_{\mathrm{DISCO}}^{\mathrm{mask}}$, we need to draw samples from the ``zero-level'' posterior, conditioned on $\xx_{\mm}$: 
\begin{equation}
p_0(\xx_{\bmm} \cond \xx_{t,\bmm},\xx_{\mm}) \propto q_0(\xx_{t,\bmm} \cond \xx_{\bmm}) p_{\text{emp}}(\xx_{\bmm} \cond \xx_{\mm})
\end{equation}
In all image modeling experiments where we employ this, we will have that $p_{\text{emp}}(\xx_{\bmm} \cond \xx_{\mm})$ will be a single Dirac-delta $\delta(\xx_{\bmm} - \xx'_{\bmm})$ where $\xx'_{\bmm}$ is the ``helper'' data point that was used to produce $\xx_{t,\bmm}$ (since $p(\mm)$ only has support on ``large enough'' masks $\mm \neq \mathbf{0}$). Therefore, sampling from $\varphi(t, \mm, \xx_{t,\bmm}, \xx_{\mm}, \xx_{\bmm}) := p(t)\,p(\mm)\,p_t(\xx_{t, \bmm}, \xx_{\mm})\,p_0(\xx_{\bmm} \cond \xx_{t,\bmm},\xx_{\mm})$ is trivial by sampling $t \sim p(t)$, $\mm \sim p(\mm)$, $\xx' \sim p^*$, $\xx_{t,\bmm} \sim q_t(\xx_{t,\bmm} \cond \xx'_{\bmm})$ and setting $\xx_{\bmm} := \xx'_{\bmm}$ and $\xx_{\mm} := \xx'_{\mm}$.

\section{Conditional Sampling vs.\ Inverse Problems}
\label{app:inverse}
A related but distinct problem is Bayesian inference in inverse problems, where one observes $\mathbf{y} \sim p(\mathbf{y} \cond \xx)$ through a known, typically noisy forward model and samples $p_{\btheta}(\xx \cond \mathbf{y}) \propto p(\mathbf{y} \cond \xx) \, p_{\btheta}(\xx)$ using a pretrained diffusion or score-based prior, e.g., via DPS \citep{chung2023diffusion}, plug-and-play diffusion models (PnP-DM) \citep{wu2024principled}, policy-gradient guidance (DPG) \citep{tang2024solving}, or annealed Langevin methods with guarantees (APMC) \citep{sun2024provable}.
Conditioning on clean coordinates is a special case only with the noise-free likelihood $p(\mathbf{y} \cond \xx) = \delta(\mathbf{y} - \Omega \xx)$, for which the posterior is singular on the ambient space $\mathbb{R}^D$; DISCO avoids this by sampling directly in the unobserved coordinates with the conditional score \eqref{eq:conditional_score}.
These methods are therefore complementary: they address how to use a given prior under an explicit likelihood, whereas DISCO addresses how to \emph{train} a score-based joint model whose arbitrary clean-coordinate conditionals are directly and coherently accessible.
In particular, our \emph{Gradient Guidance} baseline is the DPS-style likelihood-gradient approximation specialized to clean-coordinate conditioning, and we additionally report DPS itself in Table~\ref{tab:inpainting-results}.

\section{DISCO Training Algorithm}
\label{app:algorithm}

\begin{algorithm}[t]
\caption{One DISCO training step}
\label{alg:disco}

\begin{algorithmic}[1]
\Require dataset $\dat = \{\xx^{(i)}\}_{i=1}^N$, noise scale $\sigma(\cdot)$, $p(t)$, $\lambda(t)$, score network $s_{\btheta}$, batch size $B$
\For{$b = 1, \dots, B$}
  \State $t_b \sim p(t)$, \quad helper sample $\xx'_b \sim \mathrm{Unif}(\dat)$, \quad $\ee_b \sim \mathcal{N}(\mathbf{0}, I)$
  \State $\xx_{t,b} \gets \xx'_b + \sigma(t_b)\, \ee_b$ \Comment{exact sample from $p_{t_b}$}
  \State $\ell_{bi} \gets -\|\xx_{t,b} - \xx^{(i)}\|_2^2 / (2\sigma(0)^2)$ for all $\xx^{(i)} \in \dat$ \Comment{or over the mini-batch $\{\xx'_1, \dots, \xx'_B\}$}
  \State $i_b \sim \mathrm{Categorical}(\mathrm{softmax}(\ell_{b\cdot}))$, \quad $\xx_b \gets \xx^{(i_b)}$ \Comment{$\xx_b \sim p_0(\xx \cond \xx_{t,b})$}
\EndFor
\State $\hat{\mathcal{L}}(\btheta) \gets \frac{1}{B} \sum_{b=1}^B \lambda(t_b)\, \big\| (\xx_b - \xx_{t,b}) / \sigma(0)^2 - s_{\btheta}(\xx_{t,b}) \big\|_2^2$
\State update $\btheta$ with a gradient step on $\hat{\mathcal{L}}(\btheta)$
\end{algorithmic}

\end{algorithm}

Algorithm~\ref{alg:disco} summarizes one training step of DISCO with a directly parameterized score network; for an energy-based model, $s_{\btheta}(\xx) = -\nabla_{\xx} E_{\btheta}(\xx)$. The posterior in line 4 is either computed exactly over the full dataset $\dat$ (low-dimensional experiments) or approximated over the current mini-batch (image experiments). For $\mathcal{L}_{\mathrm{DISCO}}^{\mathrm{mask}}$, one additionally draws a mask $\mm \sim p(\mm)$ per sample, noises only the coordinates $\bmm$, uses the helper sample $\xx'_{\bmm}$ as target (Appendix~\ref{app:knn}), and adds the resulting loss with weight $\gamma$.

\pagebreak
\section{Ablation: Mini-Batch Posterior Approximation}
\label{app:minibatch_ablation}
In our image experiments, we sample the zero-level posterior $p_0(\xx \cond \xx_t)$ over a mini-batch instead of the full dataset (Appendix~\ref{app:knn}). To isolate the effect of this approximation, we repeat the low-dimensional experiments of Section~\ref{sec:lowD} (Table~\ref{tab:w1_results_2d}), but compute the posterior mass function over mini-batches of different sizes; all other settings are unchanged. Tables~\ref{tab:minibatch_fit} and~\ref{tab:minibatch_inference} report model fit and inference quality, respectively, with the same metrics as Table~\ref{tab:w1_results_2d} (mean $\pm\, 2\sigma$).

The effect on joint model fit is dataset-dependent: Checkerboard is essentially unchanged, Moons degrades only mildly, whereas Rings benefits noticeably from larger batches and exact posterior sampling. Since DISCO targets the conditional of the learned \emph{model}, inference quality remains uniformly strong across all batch sizes.

\begin{table}[ht]
\centering
\caption{\textbf{Mini-batch posterior, model fit:} $W_1(p_0(\xx), p_{\btheta}(\xx))$ for DISCO trained with the zero-level posterior computed over mini-batches of the given size, or exactly over the full dataset (as in Table~\ref{tab:w1_results_2d}).}
\label{tab:minibatch_fit}

\begin{tabular}{rccc}
\toprule
Batch size & Moons & Checkerboard & Rings \\
\midrule
64 & $0.085 \pm 0.031$ & $0.038 \pm 0.011$ & $0.340 \pm 0.040$ \\
128 & $0.081 \pm 0.018$ & $0.032 \pm 0.008$ & $0.230 \pm 0.039$ \\
256 & $0.111 \pm 0.046$ & $0.039 \pm 0.011$ & $0.170 \pm 0.030$ \\
512 & $0.093 \pm 0.039$ & $0.032 \pm 0.007$ & $0.117 \pm 0.021$ \\
Full dataset & $0.068 \pm 0.022$ & $0.029 \pm 0.008$ & $0.066 \pm 0.026$ \\
\bottomrule
\end{tabular}

\end{table}

\begin{table}[ht]
\centering
\caption{\textbf{Mini-batch posterior, inference quality:} $\mathbb{E}_{p_0(x_1)} [ W_1(p_{\btheta}(x_2 \cond x_1), \hat{p}_{\btheta}(x_2 \cond x_1)) ]$ for the same models as in Table~\ref{tab:minibatch_fit}, with $p_{\btheta}$ the exact model conditional (via rejection sampling) and $\hat{p}_{\btheta}$ obtained by tempered SMC on the conditional score \eqref{eq:conditional_score}.}
\label{tab:minibatch_inference}

\begin{tabular}{rccc}
\toprule
Batch size & Moons & Checkerboard & Rings \\
\midrule
64 & $0.023 \pm 0.024$ & $0.016 \pm 0.019$ & $0.023 \pm 0.028$ \\
128 & $0.031 \pm 0.036$ & $0.014 \pm 0.013$ & $0.018 \pm 0.020$ \\
256 & $0.027 \pm 0.029$ & $0.016 \pm 0.017$ & $0.033 \pm 0.045$ \\
512 & $0.024 \pm 0.024$ & $0.014 \pm 0.014$ & $0.031 \pm 0.029$ \\
Full dataset & $0.024 \pm 0.025$ & $0.015 \pm 0.012$ & $0.042 \pm 0.040$ \\
\bottomrule
\end{tabular}

\end{table}

\section{EDM-Based DISCO Training}
\label{app:edm_disco}

We use the \emph{EDM} implementation \citep{edm}, which parameterizes a \emph{denoising network} $D_{\btheta}(\xx)$, from which the corresponding score network is obtained as
\begin{equation}
   s_{\btheta}(\xx) := \frac{D_{\btheta}(\xx) - \xx}{\sigma(0)^2}.
\end{equation}
Since $\nabla_{\xx_t} \log q_0(\xx_t \cond \xx) = (\xx - \xx_t)/\sigma(0)^2$, it follows that $\mathcal{L}_{\mathrm{DISCO}}$ simplifies to 
\begin{equation}
\mathcal{L}_{\mathrm{DISCO}}(\btheta) = \sigma(0)^{-4} \ \mathbb{E}_{q(t,\xx, \xx_t)} \left[\lambda(t) \|\xx - D_{\btheta}(\xx_t) \|^2_2 \right]
\end{equation}
where we drop $\sigma(0)^{-4}$ because it is a constant factor w.r.t.\ $\btheta$.
\citet{edm} model their time-dependent denoiser as 
\begin{equation}
    D_{\btheta}(\xx, t) := c_{\text{skip}}(t) \xx + c_{\text{out}}(t) F_{\btheta}(c_\text{in}(t)\xx, c_{\text{noise}}(t))
\end{equation}
where $F_{\btheta}(\cdot, \cdot)$ is the direct output of the neural network, and $c_{\text{skip}}$, $c_{\text{out}}$, $c_{\text{in}}$, $c_{\text{noise}}$ are scalar-valued functions.
Since we model a time-independent denoiser, we do not use $c_{\text{noise}}$ and let the network $F_{\btheta}$ predict both $c_{\text{skip}}$ and $c_{\text{out}}$ via a linear layer with a sigmoid activation at the output.
We train $D_{\btheta}(\xx)$ by minimizing $\mathcal{L}_{\mathrm{DISCO}}^{\mathrm{mask}}$ with $\gamma = 0.5$, using the mini-batch approximation to sample from the posterior $p_0(\xx \cond \xx_t)$ (see Appendix~\ref{app:knn}). Our mask distribution $p(\mm)$ includes randomly sized patches and factorized Bernoulli distributions for the mask variables.

\section{Experimental Details}
\label{app:experimental_details}

\paragraph{Computing Infrastructure.} All experiments ran on GPU nodes of SLURM clusters. The low-dimensional models were trained and evaluated on single NVIDIA Quadro RTX 8000 or L40 GPUs. Each image model (DISCO and \emph{EDM Masked}) was trained data-parallel on a single node with $4$--$8$ NVIDIA Quadro RTX 8000 or L40 GPUs. Sampling and evaluation (FID, inpainting, and the resampling consistency test) ran on NVIDIA Quadro RTX 8000, L40, H100 and B200 GPUs.

\paragraph{Assets and Licenses.} We use the following existing assets.
\emph{Datasets:} CIFAR-10 \citep{krizhevsky2009learning}, distributed by its authors for research use without an explicit license; FFHQ \citep{ffhq}, whose compilation is licensed under CC BY-NC-SA 4.0 by NVIDIA, with the individual images under CC BY 2.0, CC BY-NC 2.0, Public Domain Mark 1.0, CC0 1.0, or U.S.\ Government Works licenses; the two-moons data is generated with scikit-learn \citep{scikit-learn} (BSD-3-Clause), and all other low-dimensional data is synthetic.
\emph{Code and models:} the EDM code base and the pretrained EDM checkpoints \citep{edm} (CC BY-NC-SA 4.0), including the Inception-v3 network used for FID; the score SDE code base \citep{song2020sde} (Apache 2.0); LPIPS \citep{lpips} (BSD-2-Clause); and the official TDS implementation \citep{wu2023practical}, which is publicly available without an explicit license and which we only used to evaluate the TDS baseline, without redistributing it.
Our code is derived from EDM and the score SDE code base and is released under their respective licenses (CC BY-NC-SA 4.0 for the image experiments, Apache 2.0 for the low-dimensional experiments).

\subsection{Low-Dimensional Setting}

\paragraph{DISCO Samples.} 
As discussed in Section \ref{sec:score_based_modeling}, we might use \emph{unconstrained} or \emph{energy-based} score models for training DISCO, which influences our options for sampling.
For unconstrained scores, we might use Unadjusted Langevin Algorithm (ULA) to draw samples from our model. 
However, since it is well known that Langevin algorithms suffer from slow mixing times if the target distribution is multimodal, we employ \emph{tempering} strategies \citep{neal1996sampling} by considering a sequence of distributions $\{ p_{\beta_i} \}_{i=0}^n$ with  
\begin{equation}
    p_{\beta_i}(\xx) \propto p_{\btheta}(\xx)^{\beta_i}
\end{equation}
where $0 = \beta_0 < \dots < \beta_n =1$ is a schedule of \emph{inverse} temperature parameters. 
As $\beta \to 0$, $p_{\beta}$ approaches a uniform distribution, and as $\beta \to 1$, we recover the original model $p_{\btheta}$. 
Tempering simply scales the score, i.e., $\nabla_{\xx} \log p_{\beta}(\xx) = \beta \nabla_{\xx} \log p_{\btheta}(\xx)$.
In the same way, we can also temper any \emph{conditional} distribution of $p_{\btheta}$ given by \eqref{eq:conditional_score}.

For sampling from the DISCO model, we use \emph{BlackJAX}~\citep{cabezas2024blackjax} and apply tempered sequential Monte Carlo (SMC) \citep{naesseth2019elements, doucet2001sequential, chopin2020introduction, del2006sequential} with an adaptive schedule for the inverse temperatures \( \beta_i \), using an ESS target of $0.75$. We perform systematic resampling after $2$ Hamiltonian Monte Carlo (HMC) steps, using $3$ leapfrog integration steps each. Since $\beta_0 = 0$, we initialize the SMC sampler with the uniform distribution between $-2.5$ and $2.5$.

\paragraph{Wasserstein-1 Distance.} To evaluate model fit, we sample from the model $10,000$ times, sample from $p_0$ $10,000$ times, and compute (discrete) $W_1$ distance between them. We repeat this $30$ times and report mean $\pm$ two times the standard deviation.
To evaluate inference quality, we draw $10^6$ samples from the (joint) model. We then sample $30$ times from $p_0(x_1)$ and for each $x_1$, we use rejection sampling (with $\epsilon = 0.02$) to slice out the conditional samples from the joint samples. We then sample from the conditional model (e.g. using a heuristic, or DISCO via the same tempered SMC sampler) and again compute $W_1$ between these sets. Finally, we report the mean $W_1$ distance $\pm$ two times the standard deviation.

\paragraph{Diffusion Model Samples.} 
To sample from the (joint) diffusion model, we use the ancestral sampling scheme with $100$ steps. This is roughly equivalent in terms of computational cost to sampling from the DISCO model. The conditional heuristics use the same $100$ ancestral steps. For DPS \citep{chung2023diffusion}, we selected the step size $\zeta$ per dataset from $\{0.01, 0.03, 0.1, 0.3, 1, 3, 10\}$ by inference quality on $10$ held-out conditioning values ($\zeta = 0.3$ for Moons and $\zeta = 0.1$ for Checkerboard and Rings).

\paragraph{Weighting Function.} In all low-dimensional experiments, we let $\log(t) \sim \mathcal{U}(\log(\sigma(0)), \log(\sigma(T)))$ and use $\lambda(t) = 1$ for all $t$.

\paragraph{Network Architecture.} The network architectures of the diffusion models and DISCO models are identical, except that the diffusion models receive the noise level $\sigma(t)$ as input, while the DISCO models do not. In the former case, we use a simple positional embedding for $\sigma(t)$, which we concatenate to the input. Moreover, following \citet{tancik2020fourier}, we also use the same positional embedding for each coordinate in the input $\xx$.
The remainder of the MLP consists of $4$ blocks (with residual connections), where each block contains $2$ affine layers followed by leaky ReLU activations, and normalization layers at the start of each block, after the first affine layer (\texttt{InstanceNorm++} introduced in \citet{song2019gen}). 
All affine layers in these blocks are maps $\mathbb{R}^K \to \mathbb{R}^K$, where we choose $K = 128$ when $D=2$.
The final block is followed by the same normalization and activation layers, and a final affine layer mapping from $\mathbb{R}^K$ to $\mathbb{R}^D$.
When parameterizing the score directly, we use the output of the final hidden layer $\mathbf{z}$ as our score approximation. When building an energy-based model, we follow \citet{du2023reduce} and compute the energy $E_{\btheta}$ as $-\|\mathbf{z}\|^2_2$.

\subsection{High-Dimensional Setting}

\paragraph{Training.} To train both \emph{DISCO} and the \emph{EDM Masked} model, we follow the EDM repository\footnote{\url{https://github.com/NVlabs/edm}} and use the same training hyperparameters of \texttt{edm-cifar10-32x32-uncond-vp} and \texttt{edm-ffhq-64x64-uncond-vp}, respectively---except for the following changes: We set the Adam hyperparameter $\beta_2 = 0.95$ for DISCO ($\beta_2 = 0.999$ for \emph{EDM Masked}, as in EDM) and use a batch size of $B = 448$ (CIFAR) and $B = 224$ (FFHQ). Each batch is constructed by sampling $B/2$ clean images, and using the same images for both the unmasked and the masked loss (sampling fresh masks every time).

\paragraph{Checkpoints.} All post-hoc heuristics (Replacement, Gradient Guidance, DPS, RePaint, TDS) and the EDM column of Table~\ref{tab:fid} use the official pretrained EDM checkpoints \texttt{edm-cifar10-32x32-uncond-vp} and \texttt{edm-ffhq-64x64-uncond-vp}. \emph{EDM Masked} uses the snapshot after $100.4$M (CIFAR-10) and $100.1$M (FFHQ-64) training images in both tables. We trained DISCO for $200$M (CIFAR-10) and $187.7$M (FFHQ-64) images, saving snapshots every $5$--$10$M images. For unconditional generation (Table~\ref{tab:fid}), we select the snapshot with the lowest FID: $135.5$M images for CIFAR-10 (50k samples, over $19$ snapshots) and $160.2$M images for FFHQ-64 (8{,}192 samples, over the snapshots between $120.2$M and $187.7$M images). For inpainting (Table~\ref{tab:inpainting-results}), we use the final snapshots ($200$M and $187.7$M images), which scored better in the sampler search described below.

\paragraph{Sampler Hyperparameters.} EDM and \emph{EDM Masked} use the second-order Heun sampler of \citet{edm} with its default settings ($\rho = 7$; $18$ steps, i.e., NFE $= 35$, for CIFAR-10 and $40$ steps, i.e., NFE $= 79$, for FFHQ-64); for inpainting, \emph{EDM Masked} additionally uses $S_\text{churn} = 80$, $S_\text{min} = 0.01$, $S_\text{max} = 80$ and $S_\text{noise} = 1.007$, and on FFHQ-64 the same $300$ steps as DISCO. For DISCO, we tuned the hyperparameters of the same (stochastic) Heun sampler---number of steps, $\rho$, and $S_\text{churn}$, $S_\text{min}$, $S_\text{max}$, $S_\text{noise}$---by a search scored with FID on 8{,}192 samples (unconditional) and with mean LPIPS minus mean SSIM over all six masks on $512$ images per mask (inpainting).

\paragraph{Unconditional Sampling.} We follow \citet{edm} and their repository and use the same sampling settings for both CIFAR and FFHQ, except for DISCO, where we use the stochastic Heun sampler with $40$ steps, $\rho = 2.1$ and $S_\text{churn} = 40$ (CIFAR-10), and $\rho = 200$ and $S_\text{churn} = 2$ (FFHQ-64), with $S_\text{min} = 0.05$, $S_\text{max} = 80$ and $S_\text{noise} = 1.005$.

\paragraph{Inpainting.} For DISCO, we use the stochastic Heun sampler with $S_\text{churn} = 80$, $S_\text{min} = 0.01$, $S_\text{max} = 80$ and $S_\text{noise} = 1.007$, with $18$ steps (CIFAR-10) and $300$ steps (FFHQ-64), and a mask-dependent $\rho$.
On CIFAR-10, we use $\rho = 1.5$ for \emph{Narrow} and \emph{Wide}, $\rho = 1.8$ for \emph{Altern. Lines} and \emph{Super-Resolve 2x}, $\rho = 7$ for \emph{Expand}, and $\rho = 10^5$ for \emph{Half}.
On FFHQ-64, we use $\rho = 200$ for \emph{Wide}, $\rho = 10^3$ for \emph{Super-Resolve 2x}, $\rho = 10^5$ for \emph{Narrow}, \emph{Half} and \emph{Altern. Lines}, and $\rho = 7$ for \emph{Expand}.
The post-hoc heuristics use Euler--Maruyama ancestral sampling with the VP discretization of \citet{edm} and $36$ (CIFAR-10) or $80$ (FFHQ-64) steps, matching the NFE of the default Heun sampler. For TDS, we use the official implementation of \citet{wu2023practical} with $K = 4$ particles per conditioning image and the same number of steps, and evaluate the particle with the largest final importance weight. DPS \citep{chung2023diffusion} follows their Algorithm~1 (an unguided ancestral step, followed by a step of size $\zeta$ along $-\nabla_{\xx_t} \|\xx^c - \Omega \hat{\xx}_{\btheta}(\xx_t, t)\|_2$) with the same steps as the other heuristics; $\zeta = 3$ was selected from $\{0.3, 1, 3, 10, 30\}$ by LPIPS and SSIM on the first $256$ images of the \emph{Wide} and \emph{Half} masks. For all methods, the observed pixels of each sample are set to the conditioning image, both for LPIPS and SSIM and in Figure~\ref{fig:ffhq_inpainting}.
LPIPS and SSIM are computed on the first $1{,}024$ images of the CIFAR-10 test set and of the FFHQ-64 dataset, respectively; the inpainting sampler search used the first $512$ of these images.

The \emph{RePaint} heuristic uses a jump size of $1$ and $4$ resampling steps, while adjusting the outer loop iteration count such that the total number of function evaluations is the same.

\section{Resampling Consistency Test}
\label{app:gibbs}

\begin{figure}[ht]
\centering
\includegraphics[width=\linewidth]{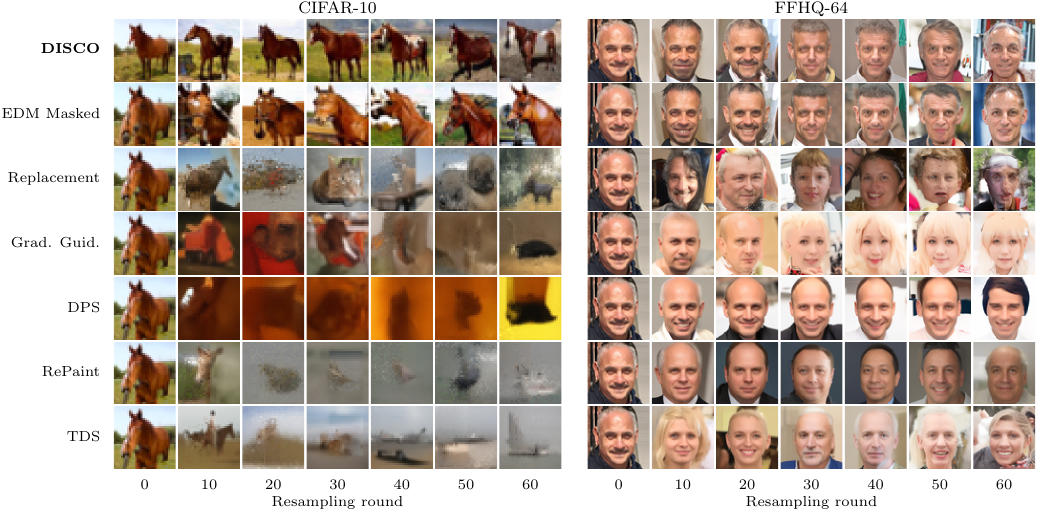}
\caption{One chain of the resampling consistency test (seed 0) per method, shown every $10$ rounds. DISCO and \emph{EDM Masked} start from their own samples, the post-hoc samplers from the same sample of the pretrained EDM. Under a consistent sampler the image content may change from round to round (the chain mixes), but its quality should not degrade.}
\label{fig:gibbs_chains}
\end{figure}

Each of $10{,}000$ chains starts from an unconditional sample of the respective model (for the post-hoc samplers use samples of the pretrained EDM) and repeats for $60$ rounds: draw a mask from the masked-training distribution (left or right half, Bernoulli$(0.5)$ pixels, or a random rectangle covering at least $25\%$ of the image), keep the observed pixels, resample the remaining ones with the method's conditional sampler, and quantize to 8 bits. All methods see identical masks; the four seeds change the starting samples, masks and sampler noise. We report FID against the dataset statistics on all $10{,}000$ chains after every round; Figure~\ref{fig:gibbs} shows the mean over the four seeds (example chains in Figure~\ref{fig:gibbs_chains}), whose standard deviations (at most $0.3$ FID for DISCO and \emph{EDM Masked}, and at most $1.1$ for all methods) are too small to be visible.
DISCO and \emph{EDM Masked} use the snapshots after $100.4$M (CIFAR-10) and $100.1$M (FFHQ-64) training images and the deterministic Heun sampler ($\rho = 7$, $S_\text{churn} = 0$, $18$ and $40$ steps); DISCO stops at its smallest training noise level $\sigma_\text{min} = \sigma(0) = 0.01$. The post-hoc samplers use the settings of Table~\ref{tab:inpainting-results}, except that TDS uses our own implementation in the EDM code base, with $K = 2$ particles per chain.

\clearpage
\section{Additional High-Dimensional Qualitative Results}
\label{app:high_dim_qualitative}

\begin{center}
\includegraphics[width=\linewidth]{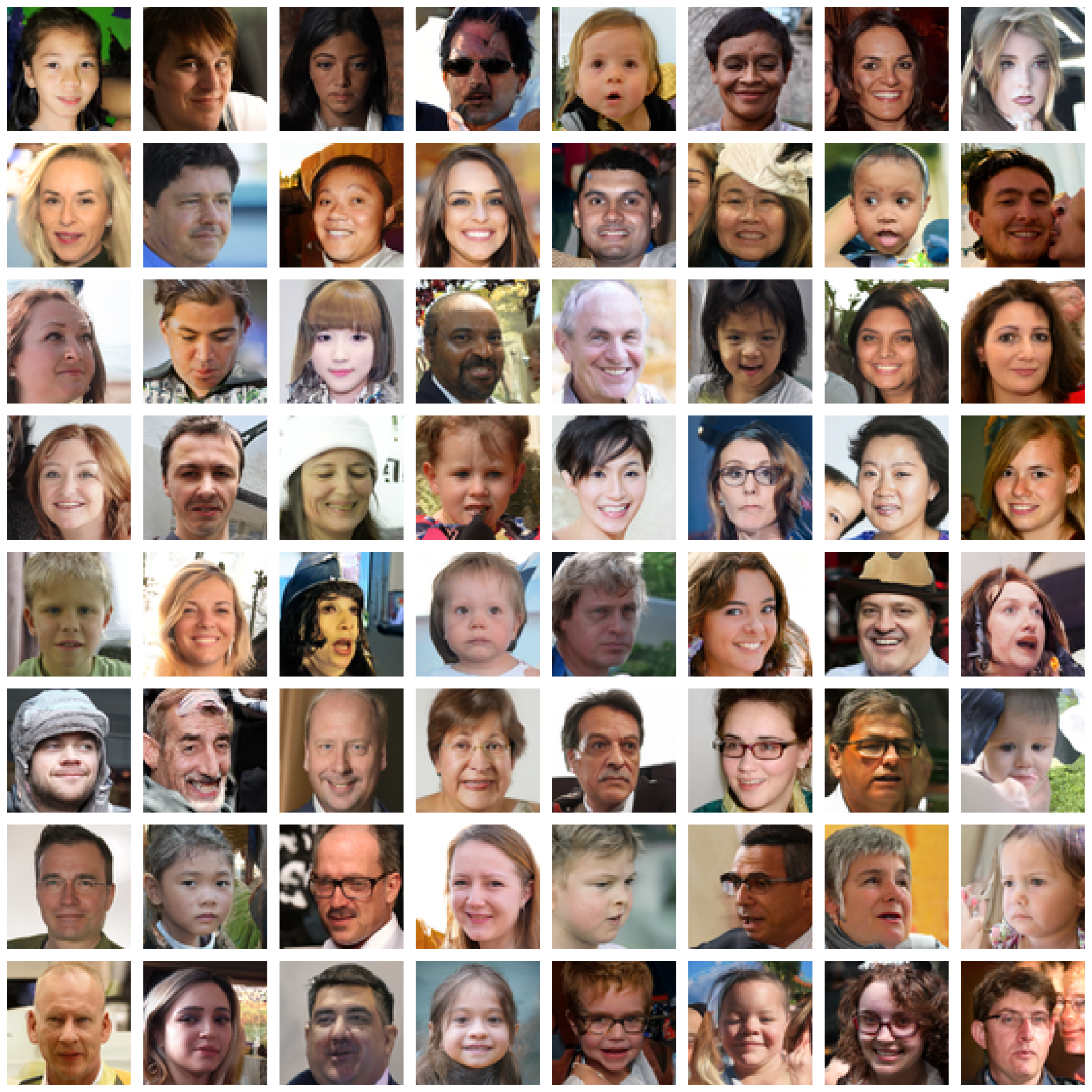}
\captionof{figure}{Unconditional samples from the DISCO model trained on FFHQ-64.}
\label{fig:ffhq_images_disco}
\end{center}
\clearpage

\begin{figure}[p]
\centering
\includegraphics[width=\linewidth]{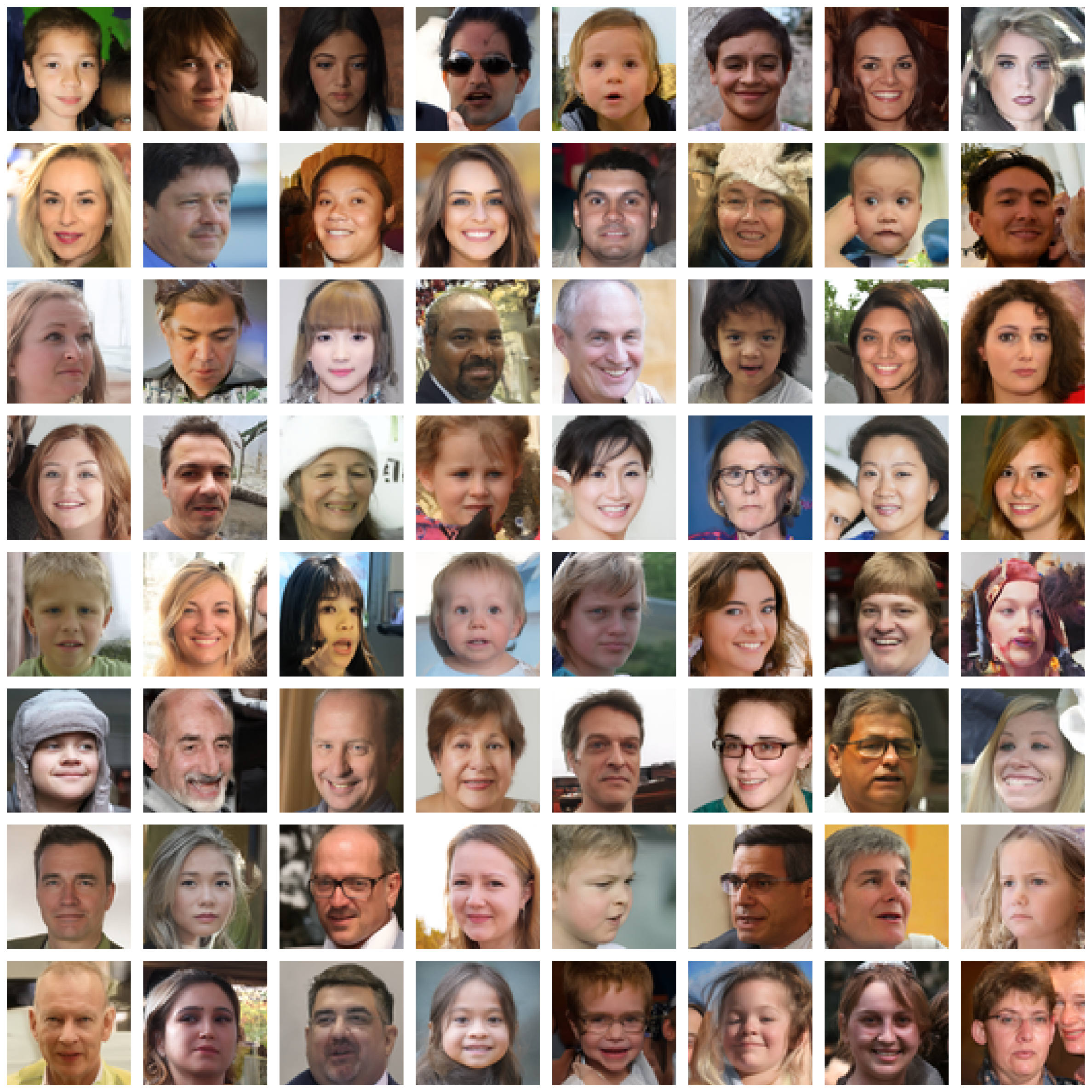}
\label{fig:ffhq_images_dm}
\caption{Unconditional samples from the diffusion model trained on FFHQ-64.}
\end{figure}

\begin{figure}[p]
\centering
\includegraphics[width=\linewidth]{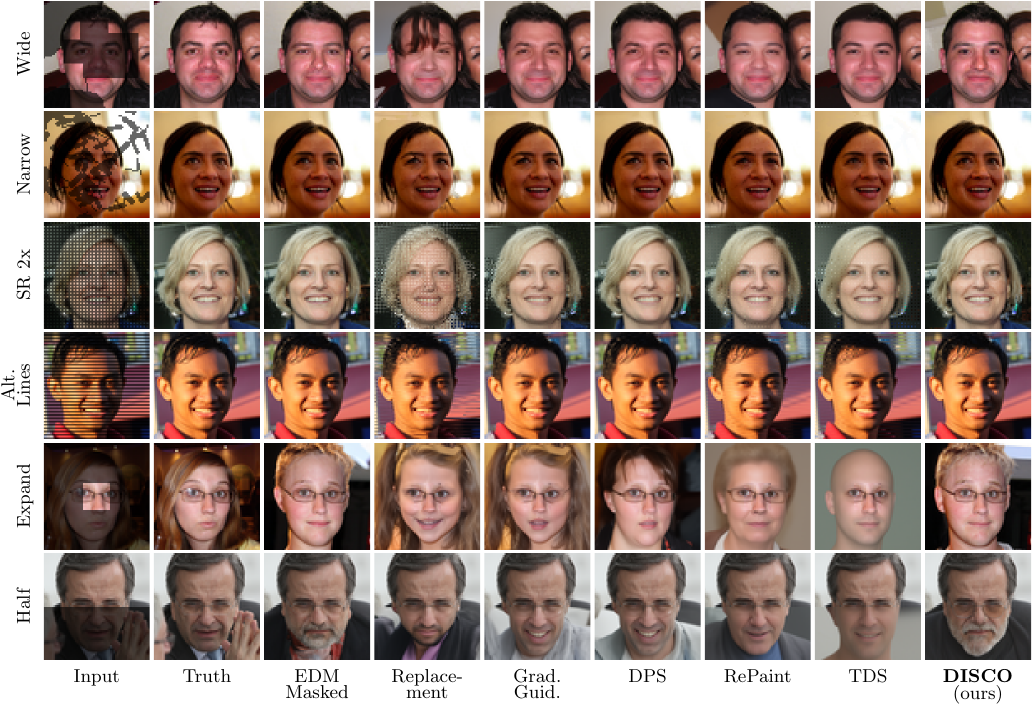}
\label{fig:inpaint_row6_2}
\caption{Qualitative inpainting results on FFHQ-64 (images 16--21 of the evaluation set, one per mask), with the samplers of Table~\ref{tab:inpainting-results}; \emph{Input} dims the unobserved pixels, \emph{Truth} shows the original image.}
\end{figure}

\begin{figure}[p]
\centering
\includegraphics[width=\linewidth]{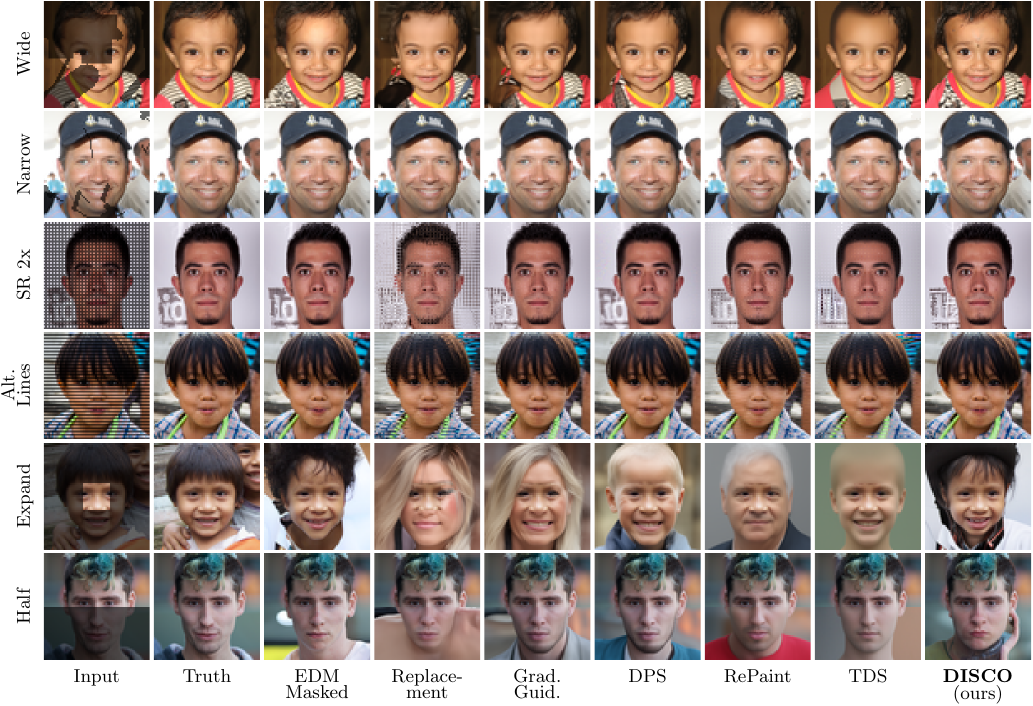}
\label{fig:inpaint_row6}
\caption{Qualitative inpainting results on FFHQ-64 (images 10--15 of the evaluation set, one per mask), with the samplers of Table~\ref{tab:inpainting-results}; \emph{Input} dims the unobserved pixels, \emph{Truth} shows the original image.}
\end{figure}

\begin{figure}[p]
\centering
\includegraphics[width=\linewidth]{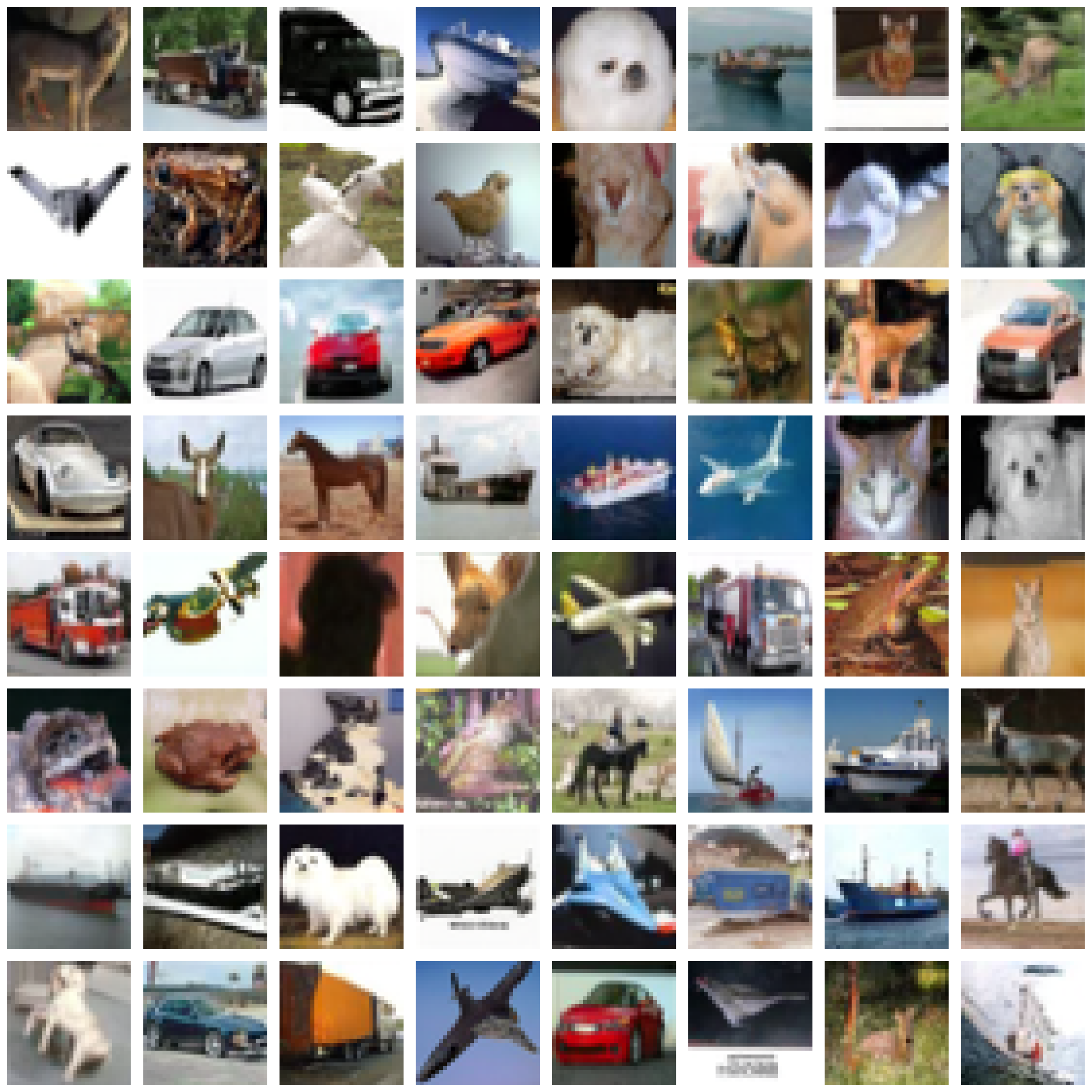}
\label{fig:cifar_images_disco}
\caption{Unconditional samples from the DISCO model trained on CIFAR-10.}
\end{figure}

\begin{figure}[p]
\centering
\includegraphics[width=\linewidth]{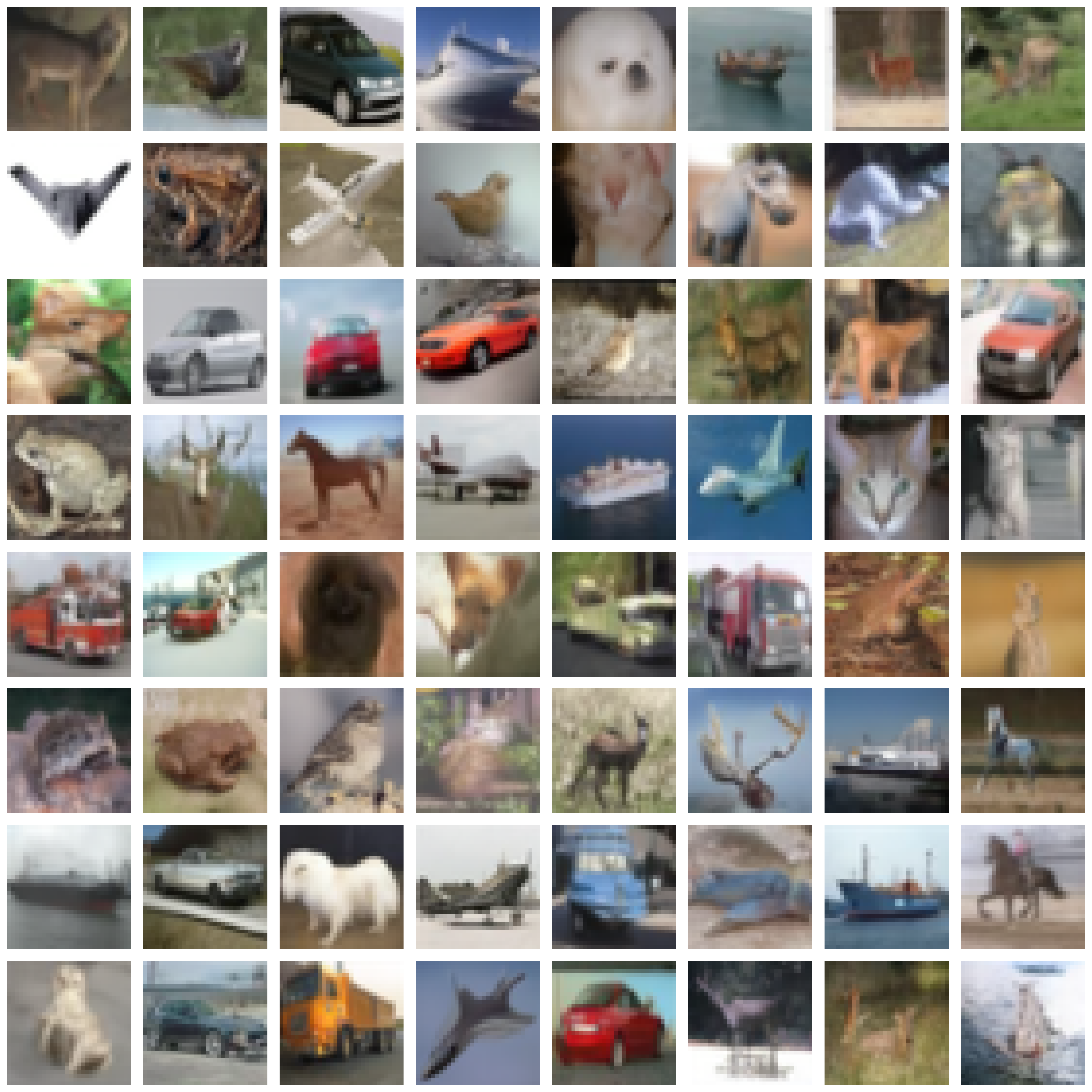}
\label{fig:cifar_images_dm}
\caption{Unconditional samples from the diffusion model trained on CIFAR-10.}
\end{figure}

\end{document}